\newcounter{myctr}

\documentclass{ws-ijhr}

\usepackage{graphicx}
\usepackage{subcaption} 

\usepackage{amssymb}
\usepackage{amsmath}
\usepackage{commath}
\usepackage{mathtools}

\usepackage{hyperref}
\usepackage{import}
\usepackage{paralist}
\usepackage{gensymb}
\usepackage{dsfont}
\usepackage{siunitx}
\usepackage[bottom]{footmisc}
\usepackage{framed}
\usepackage{balance} 
\usepackage{caption}
\usepackage[skins]{tcolorbox}
\usepackage[disable]{todonotes}

\newcommand*\diff{\mathop{}\!\mathrm{d}}

\newtcolorbox{myframe}[2][]{%
  enhanced,colback=white,colframe=black,coltitle=black,
  sharp corners,boxrule=0.4pt,left=0pt,right=0pt,top=0pt,bottom=0pt,
  fonttitle=\itshape,
  attach boxed title to top left={yshift=-0.3\baselineskip-0.8pt,xshift=2mm},
  boxed title style={tile,size=minimal,left=0.5mm,right=0.5mm,
  colback=white,before upper=\strut},
  title=#2,#1
}
\DeclareCaptionFont{mysize}{\fontsize{8}{10}\selectfont}
\begin{document}

\markboth{Romualdi et al.}{Benchmarking of DCM based architectures}

\title{A BENCHMARKING OF DCM BASED ARCHITECTURES FOR POSITION, VELOCITY AND TORQUE CONTROLLED HUMANOID ROBOTS}

\author{Giulio Romualdi $^{* \dagger}$, Stefano Dafarra $^{* \dagger}$, Yue Hu $^{\ddagger}$, \\ 
Prashanth Ramadoss $^{* \dagger}$, Francisco Javier Andrade Chavez $^{*}$, \\
Silvio Traversaro $^{*}$ and Daniele Pucci $^{*}$}

\address{$*$ Dynamic Interaction Control, Istituto Italiano di Tecnologia,\\
Genoa, Italy,\\
name.surname@iit.it}

\address{$\dagger$ DIBRIS, University of Genoa, Genoa, Italy}

\address{$\ddagger$ Intelligent Systems Research Institute, \\
National Institute of Advanced Industrial Science and Technology, Tsukuba, Japan \\
surname.name@aist.go.jp}

\maketitle

\begin{abstract}

This paper contributes towards the benchmarking of control architectures for bipedal robot locomotion. It considers architectures that are based on the Divergent Component of Motion (DCM) and composed of three main layers: \emph{trajectory optimization}, \emph{simplified model control}, and \emph{whole-body QP control} layer. While the first two layers use simplified robot models, the \emph{whole-body QP control}  layer uses a complete robot model to produce either desired positions, velocities, or torques inputs at the joint-level. 
This paper then compares two implementations of the \emph{simplified model control} layer, which are tested with position, velocity, and torque control modes for the \emph{whole-body QP control} layer. In particular, both an instantaneous and a Receding Horizon controller are presented for the \emph{simplified model control} layer. We show also that one of the proposed architectures allows the humanoid robot iCub to achieve a forward walking velocity of 0.3372 meters per second, which is the highest walking velocity achieved by the iCub  robot.
\end{abstract}


\section{Introduction}

Bipedal locomotion of humanoid robots remains an open problem despite decades of research in the subject. The complexity of the robot dynamics, the unpredictability of its surrounding environment, and the low efficiency of the robot actuation system are only few problems that complexify the achievement of robust robot locomotion. In the large variety of robot controllers for bipedal locomotion, the Divergent-Component-of-Motion (DCM) is an ubiquitous concept used for generating walking patterns. This paper presents and compares different DCM based control architectures for humanoid robot locomotion.

\par

During the DARPA Robotics Challenge, a common approach for humanoid robot control consisted in defining a hierarchical architecture composed of several layers.\cite{feng2015optimization} Each layer generates references for the layer below by processing inputs from the robot, the environment, and the outputs of the layer before. From top to bottom, these layers are here called: \emph{trajectory optimization}, \emph{simplified model control}, and \emph{whole-body quadratic programming (QP) control}.

The \emph{trajectory optimization} layer often generates  desired foothold locations by means of optimization techniques. To do so, both kinematic and dynamical robot models can be used.\cite{dai2014whole,herzog2015trajectory} When solving the optimization problem associated with the \emph{trajectory optimization} layer, computational time may be a concern especially when the robot surrounding environment is not structured. There are cases, however, where simplifying assumptions on the robot environment can be made, thus reducing the associated computational time. For instance, flat terrain allows one to view the robot as a simple unicycle,\cite{PascalHandbook,flavigne2010reactive} which enables fast solutions to the optimization problem for the walking pattern generation.\cite{8594277}
\par
The \emph{simplified model control} layer
is in charge of finding feasible center-of-mass (CoM) trajectories and it is often based on simplified dynamical models, such as the Linear Inverted Pendulum Model (LIPM)\cite{Kajita2001} and the Capture Point (CP).\cite{Pratt2006} These models have become very popular after the introduction of the Zero Moment Point (ZMP) as a contact feasibility criterion.\cite{Vukobratovic1969} %
To obtain feasible CoM trajectories, the \emph{simplified model control} layer often combines the LIPM with Model Predictive Control (MPC) techniques, also known as the Receding Horizon Control (RHC).\cite{Kajita2003,diedam2008online}
Another model that is often exploited in the \emph{simplified model control} layer is the Divergent Component of Motion (DCM).\cite{Englsberger2015}
The DCM can be viewed as the extension of the capture point (CP) to the three dimensional case under the assumption of a constant height of the Virtual Repellent Point (VRP) respect to Enhanced Centroidal Moment Pivot point (eCMP).\cite{Englsberger2015}
Attempts at loosening this latter assumption and extending the DCM to more complex models have also been presented.\cite{Hopkins2015} %

The \emph{whole-body QP control} layer generates robot  positions, velocities or torques depending on the available control modes of the underlying robot. These outputs aim at stabilizing the references generated by the layers before. It uses whole-body kinematic or dynamical models, and very often instantaneous optimization techniques: no MPC methods are here employed. Furthermore, the associated optimisation problem is often framed as an hierarchical stack-of-tasks, with strict or weighted hierarchies.\cite{Stephens2010,nava16}
\par
Recently, the scientific community has been interested in the possibility of using torque control based  algorithms to perform locomotion tasks.\cite{Stephens2010,Lee2016,Feng2015a,Kuindersma2016,koolen_ijhr} Indeed torque-controlled robots have several advantages over position or velocity controlled ones. A torque-controlled humanoid robot is, in fact, intrinsically compliant in case of external unexpected interactions, and it can be thus used to perform cooperative tasks alongside humans.\cite{Romano2018}
\par
This paper extends and encompasses our previous work \cite{8625025} and presents and compares several DCM based implementations of the above layered control architecture. 
In particular, the \emph{trajectory optimization} layer is kept fixed with a unicycle based  planner that generates desired DCM and foot  trajectories. The \emph{simplified model control} layer, instead, implements two types of controllers for the tracking of the DCM: an instantaneous and an MPC one. 
In the same layer, we also present a controller which exploits 6-axes Force Torque sensors (F/T), thus ensuring the tracking of both the CoM and the ZMP.
Finally, the \emph{whole-body QP control} implements two controllers for the tracking of the Cartesian trajectory: a kinematics-based and a dynamics-based whole body controllers.
The former uses the kinematics of the robot for generating desired joint positions/velocities. While the latter is based on the entire robot dynamics and its output are the desired joint torques.
The several combinations of the control architecture are tested on the iCub humanoid robot.\cite{Natale2017} One of the proposed implementations allows the iCub robot to reach a forward walking velocity of $\SI{0.3372}{\meter \per \second}$.
\par
The paper is organized as follows. Sec.~\ref{sec:background} introduces notation, the humanoid robot model, and some simplified models commonly used for locomotion.
Sec.~\ref{sec:architecture} describes each layer of the control
architecture, namely the trajectory optimization, the simplified model control and the whole-body QP control layer. Sec.~\ref{sec:experimental_results} presents the experimental validation of the proposed approach, and shows an explanatory table comparing the different control approaches. Finally, Sec.~\ref{sec:conclusions_and_future_work} concludes the paper.

\section{Background}
\label{sec:background}
\subsection{Notation}
\begin{itemize}
\item $I_n$ and $0_n$ denote respectively the $n \times n$ identity and zero matrices;
\item $\mathcal{I}$ denotes an inertial frame;
\item $\prescript{\mathcal{A}}{}{p}_\mathcal{C}$ is a vector that connects the origin of frame $\mathcal{A}$ and the origin of frame $\mathcal{C}$ expressed with the orientation of frame $\mathcal{A}$;
\item given $\prescript{\mathcal{A}}{}{p}_\mathcal{C}$ and $\prescript{\mathcal{B}}{}{p}_\mathcal{C}$,  $\prescript{\mathcal{A}}{}{p}_\mathcal{C} = \prescript{\mathcal{A}}{}{R}_\mathcal{B} \prescript{\mathcal{B}}{}{p}_\mathcal{C} + \prescript{\mathcal{A}}{}{p}_\mathcal{B}= \prescript{\mathcal{A}}{}{H}_\mathcal{B} \begin{bmatrix}
  \prescript{\mathcal{B}}{}{p}_\mathcal{C} ^\top & \;1
\end{bmatrix}^\top$, where $\prescript{\mathcal{A}}{}{H}_\mathcal{B}$ is the homogeneous transformations and $\prescript{\mathcal{A}}{}{R}_\mathcal{B} \in SO(3)$ is the rotation matrix; 
\todo[inline]{ST: $\prescript{\mathcal{A}}{}{H}_\mathcal{B}$ is a $4 \times 4$ matrix, while $\prescript{\mathcal{B}}{}{p}_\mathcal{C}$ is a $3 \times 1$ vector. I am afraid we need to append a $1$ at the end of of $\prescript{\mathcal{B}}{}{p}_\mathcal{C}$}
\item given  $W\in \mathfrak{so}(3)$ the \emph{vee operator} is $.^\vee : \mathfrak{so}(3) \to \mathbb{R}^3$;
\item $\prescript{\mathcal{A}}{}{\omega}_\mathcal{B} \in \mathbb{R}^3$ denotes the angular velocity between the frame $\mathcal{B}$ and the frame $\mathcal{A}$ expressed in the frame $\mathcal{A}$;
\item given $A \in \mathbb{R}^{3 \times 3}$ the \emph{skew} operator is  $\text{sk} :\mathbb{R}^{3 \times 3} \to \mathfrak{so}(3)$;%
\item the subscripts $\mathcal{T}$, $\mathcal{LF}$, $\mathcal{RF}$ and $\mathcal{C}$ indicates the frames attached to the torso, left foot, right foot and CoM;
\item henceforth, for the sake of clarity, the prescript $\mathcal{I}$ will be omitted;
\item the superscript ${.}^{ref}$ indicates a desired quantity generated by the \emph{trajectory optimization} layer;
\item the superscript ${.}^*$ indicates a desired quantity generated by the \emph{simplified model control} layer;
\item the term \emph{pose} indicates the combination of position and orientation.
\end{itemize}

\subsection{Humanoid Robot Model}
A humanoid robot is modelled as a floating base multi-body system composed of $n+1$ links connected by $n$ joints with one degree of freedom each. Since none of the robot links has an a priori pose w.r.t. the inertial frame $\mathcal{I}$, the robot configuration is completely defined by considering both the joint positions $s$ and the homogeneous transformation from the inertial frame to the robot frame (i.e. called base frame $\mathcal{B}$). 
In details, the configuration of the robot can be uniquely determined by the triplet $q = (\prescript{\mathcal{I}}{}{p}_\mathcal{B}, \prescript{\mathcal{I}}{}{R}_\mathcal{B}, s) \in  \mathbb{R}^3 \times SO(3) \times \mathbb{R}^n$.
The velocity of the floating system is represented by the triplet $ \nu = (\prescript{\mathcal{I}}{}{v}_\mathcal{B}, \prescript{\mathcal{I}}{}{\omega}_\mathcal{B}, \dot{s})$, where $\prescript{\mathcal{I}}{}{v}_\mathcal{B}$ and $\dot{s}$ are the time derivative of the position of the base and the joint positions, respectively. $\prescript{\mathcal{I}}{}{\omega}_\mathcal{B}$ is defined as
\begin{equation}
    \prescript{\mathcal{I}}{}{\omega}_\mathcal{B} = \left( \prescript{\mathcal{I}}{}{\dot{R}}_\mathcal{B} \prescript{\mathcal{I}}{}{R}_\mathcal{B} ^\top \right)^\vee.
\end{equation}
\todo[inline]{$\prescript{\mathcal{I}}{}{\omega}_\mathcal{B}$ and $\prescript{\mathcal{I}}{}{v}_\mathcal{B}$ are not defined. While the $\prescript{\mathcal{I}}{}{\omega}_\mathcal{B}$ definition can understood for the context, $\prescript{\mathcal{I}}{}{\omega}_\mathcal{B}$ is more problematic, as it is not defined, and the fact that you refer as it as a "twist" may induce the reader in thinking that you are referring to the linear part of the inertial (i.e. right-trivialized) 6D velocity.}

\par
Given a frame attached to a link of the floating base system, its position and orientation w.r.t. the inertial frame is uniquely identified by a homogeneous transformation, $\prescript{\mathcal{I}}{}{H}_\mathcal{A} \in SE(3)$.
\par 
Similarly, the frame velocity w.r.t. the inertial frame is uniquely identified by the twist $\prescript{}{}{\mathrm{v}}_\mathcal{A} = \begin{bmatrix} \prescript{}{}{v}_\mathcal{B} ^ \top & \prescript{}{}{\omega}_\mathcal{B}^ \top \end{bmatrix}^ \top$. The function that maps $\nu$ to the twist $\prescript{}{}{\mathrm{v}}_\mathcal{A}$ is linear and its matrix representation is the well known Jacobian matrix $J_\mathcal{A}(q)$:
\begin{equation}
    \label{eq:frame_velocity}
    \prescript{}{}{\mathrm{v}}_\mathcal{A} = J_\mathcal{A}(q) \nu.
\end{equation}
For a floating base system the Jacobian can be split into two sub-matrices. One multiplies the base velocity while the other the joint velocities.
\par
Clearly, using \eqref{eq:frame_velocity}, the frame acceleration is given by:
\begin{equation}
    \prescript{}{}{\dot{\mathrm{v}}}_\mathcal{A} = J_\mathcal{A}(q) \dot{\nu} + \dot{J}_\mathcal{A}(q) \nu.
\end{equation}
The dynamics of the floating base system can be described by the Euler-Poincar\'e equation:\cite{Marsden2010}
\begin{equation}
\label{eq:robot_dynamic}
    M(q) \dot{\nu} + C(q, \nu) \nu + G(q) = B \tau + \sum_{k = 1}^{n_c} J^\top_{\mathcal{C}_k}(q) f_k,
\end{equation}
where, on the left hand side, $M(q) \in \mathbb{R} ^{(n + 6) \times(n + 6)}$ represents the mass matrix, $C(q, \nu) \in \mathbb{R} ^{(n + 6) \times(n + 6)}$ is the Coriolis and the centrifugal term, $G(q) \in \mathbb{R} ^{n + 6}$ is the gravity vector. On the right-hand side of \eqref{eq:robot_dynamic}, $B$ is a selector matrix, $\tau \in \mathbb{R}^n$ is the vector containing the joint torques and $f_k \in \mathbb{R}^6$ is a vector containing the coordinates of the contact wrench.
$n_c$ indicates the number of contact wrenches. Henceforth, we assume that at least one of the link is in contact with the environment, i.e. $n_c \ge 1$.
\par
Finally, let us also recall the concept of local and global Zero Moment Point (ZMP).\cite{Vukobratovic1969} The interaction between the robot and the environment is modeled as infinitesimal forces acting on the surface of the link in contact with the environment. The effect of the contact forces can be represented with an equivalent wrench $f = \begin{bmatrix} f_l^\top & f_a^\top \end{bmatrix}^\top$.
$f_l \in \mathbb{R}^3$ is the linear part of the wrench while $f_a \in \mathbb{R}^3$ is the torque.
Under the hypotheses of the Poinsot's theorem,\cite{Murray1994} it is easy to show that every wrench applied at a point on a link is equivalent to another wrench applied to a body-fixed point\cite{Featherstone2014} whose linear part has the same direction and magnitude of the original wrench while the angular part is parallel to the linear force.
If the point is placed on the surface that is in contact with the environment, then the point defines the local ZMP. Given a contact wrench $f = \begin{bmatrix} f_l ^ \top & f_a ^ \top \end{bmatrix} ^\top$ expressed in body frame, the local ZMP, if defined, is given by:\cite{Nori2015} $
    \prescript{l}{}{r}_{zmp} = 
    \begin{bmatrix}
    -\frac{f_{a_y}}{f_{l_z}} &  \frac{f_{a_x}}{f_{l_z}}
    \end{bmatrix} ^ \top$.
\par
 When one or more links of the floating base system are in contact with the environment and the contact surfaces belong to the same plane, the global ZMP can be defined:
\begin{equation}
    \prescript{g}{}{r}_{zmp} = \sum _ {k = 1} ^ {n_c} \prescript{l}{}{r}_{zmp} ^k \frac{f^k_{l_z}}{f_{l_z}}.
\end{equation}
where $f_{l_z}$ is the sum of all the contact forces acting along the z axis.
\par
For sake of simplicity, henceforth, the superscript $\prescript{g}{}{}$ of $\prescript{g}{}{r}_{zmp}$ will be dropped and the global ZMP will be indicated as ZMP, $r^{zmp} \in \mathbb{R}^2$.

\subsection{Simplified models}
\label{sec:simplified_models}
Consider a humanoid robot walking. Assume that the height of the CoM with respect to the stance foot is constant, and that the rate of change of the Centroidal Angular Momentum is equal to zero.\cite{Orin2013} Then, the motion of the robot can be approximated using the well known \emph{Linear inverted pendulum model} (LIPM).\cite{Kajita2001}
By definition of LIPM, the CoM trajectory belongs to a horizontal plane with a constant height $z_0$.
The simplified CoM dynamics is given by:\cite{Kajita2001}
\begin{equation}
  \label{eq:3d-lipm}
  \dot{v}_{\mathcal{C}} = \frac{1}{b^2}(p_\mathcal{C} - r_{zmp}),
\end{equation}
where $p_\mathcal{C} \in \mathbb{R}^2$ and $v_\mathcal{C} \in \mathbb{R}^2$ represent the position and the velocity of the CoM projected on the walking surface. $b$ is the pendulum time constant, i.e. $b = \sqrt{z_0/g}$ where $g$ is the gravity constant.
\par
Analogously, one can define the Divergent Component of Motion (DCM) as:\cite{Englsberger2015}
\begin{equation}
    \label{eq:dcm}
\xi = p_\mathcal{C} +  b v_\mathcal{C}.
\end{equation}
Using \eqref{eq:3d-lipm} and \eqref{eq:dcm}, the DCM time derivative is given by:
\begin{equation}
  \label{eq:dcm_dynamics}
  \dot{\xi} = \frac{1}{b}(\xi - r_{zmp}).
\end{equation}
By choosing the DCM as state variable, \eqref{eq:3d-lipm} can be decomposed into two parts:
\begin{equation}
  \label{eq:simplified-model}
\begin{bmatrix}
  v_\mathcal{C}\\
  \dot{\xi}
\end{bmatrix} = \frac{1}{b}
\begin{bmatrix}
  -I_2 & I_2\\
  0_2 & I_2
\end{bmatrix}
\begin{bmatrix}
  p_\mathcal{C}\\
  \xi
\end{bmatrix}
- \frac{1}{b}
\begin{bmatrix}
  0_2\\
  I_2
\end{bmatrix}
r_{zmp}.
\end{equation}
When performing the state space decomposition, it can be easily show that the CoM dynamics has a strictly negative real part eigenvalue, while the DCM dynamics has a strictly positive real part eigenvalue.

\section{Architecture}
\label{sec:architecture}

\begin{figure}[!t]
\centering
\includegraphics[width=\textwidth]{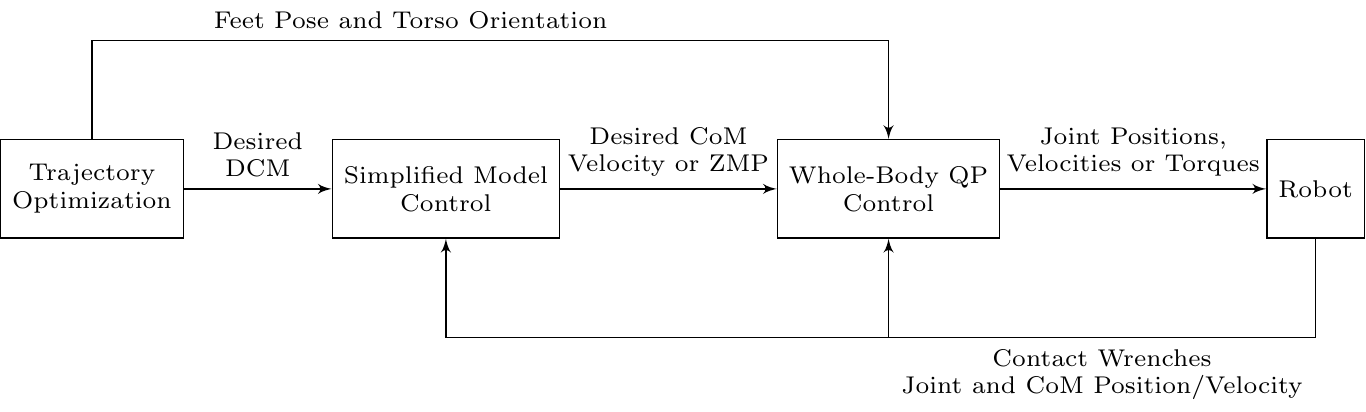}
\caption{The control architecture is composed of three layers: the \emph{trajectory optimization}, the \emph{simplified model control}, and the \emph{whole-body QP control}.
\label{fig:controller_architecture}}
\end{figure}

This section describes the component of the control architecture presented in Fig.~\ref{fig:controller_architecture}, that is investigated in this paper. The control architecture is composed of three main layers, namely the \emph{trajectory optimization}, the \emph{simplified model control} and the \emph{whole-body QP control} layers.
\par
The goal of the trajectory optimization layer is to generate the desired feet trajectory and also the desired DCM trajectory. The simplified model control layer is in charge to ensure the tracking of the desired DCM, CoM and ZMP trajectories. Lastly, the main purpose of the whole-body QP control layer is to exploit the entire model of the robot for guaranteeing the tracking of the desired cartesian trajectories.
\par
Even if the first two layers have been presented in \cite{8625025}, for the sake of completeness we recall them below.

\subsection{Trajectory optimization layer}
\label{sub:Trajectory_optimization}
The trajectory optimization layer objective is to evaluate the desired feet and DCM trajectories. 
Prior to trajectory generation, however, footsteps positions have to be planned. The humanoid robot is approximated as a unicycle, and the feet are represented by the unicycle wheels.\cite{8594277} By sampling the continuous unicycle trajectory, it is possible to associate each unicycle pose to a time instant. This time instant can be considered as the time in which the swing foot impacts the ground  $t_{imp}$. Once the impact time is defined, we decide to use it as a conditional variable in order to plan the footsteps that are feasible for the robot. Too fast/slow step duration and too long/short step length are avoided.
Once the footsteps are planned, the feet trajectory is evaluated by cubic spline interpolation.
\par
The footsteps position is also used to plan the desired DCM trajectory. In particular, the DCM is chosen so as to satisfy the following time evolution:
\begin{equation}
\label{eq:dcm_solution_ios}
\xi_i(t) = r^{zmp}_i + e ^{\frac{t}{b}} (\xi_i^{ios} - r^{zmp}_i),
\end{equation}
where $i$ indicates the $i$-th steps, $\xi_i^{ios}$ is the initial position of the DCM. $r^{zmp}_i$ is placed on the center of the stance foot while $t$ has to belong to the step domain $t \in [0, \; t^{step}_i]$.
Assuming that the final position of the DCM $\xi_{N-1}^{eos}$ coincides with the ZMP at last step, \eqref{eq:dcm_solution_ios} can be used to define the following recursive algorithm for evaluating the DCM trajectory:\cite{Englsberger2014}
\begin{equation}
\label{eq:dcm_evaluation_ss}
\begin{cases}
\xi_{N-1}^{eos} = r_{N}^{zmp} \\
\xi_{i-1}^{eos} = \xi_{i}^{ios} = r_{i}^{zmp} + e^{-\frac{t^{step}_i}{b}} (\xi_i^{eos} - r_i^{zmp}) \\
\xi_i(t) = r^{zmp}_i + e ^{\frac{t}{b}} (\xi_i^{ios} - r^{zmp}_i).
\end{cases}
\end{equation}
The presented DCM planner has the main limitation of taking into account single support phases only. Indeed, by considering instantaneous transitions between two consecutive single support phases, the ZMP reference is discontinuous. This leads to the discontinuity of the external contact wrenches and consequentially of the desired joint torques.
The development of a DCM trajectory generator that handles non-instantaneous transitions between two single support phases becomes pivotal.\cite{Englsberger2014,Englsberger2019}
To address this issue, we implemented the solution proposed by Englsberger.\cite{Englsberger2014} In details, in order to guarantee a continuous ZMP trajectory, the desired DCM trajectory must belong at least to $C^1$ class. This can be easily guaranteed by smoothing two consecutive single support DCM trajectory by means of a third order polynomial function.
Its coefficients are chosen in order to satisfy the boundaries conditions, i.e. initial and final DCM position and velocity. 

\subsection{Simplified Model Control Layer}
The main goal of the simplified model control layer is to implement a control law that stabilizes the unstable DCM dynamics \eqref{eq:dcm_dynamics}. 
The stabilization problem is faced by developing two different controllers: an \emph{instantaneous} and a \emph{predictive} one.

\subsubsection{DCM instantaneous control}
\label{sec:instantaneous-control}
To guarantee the tracking of the desired DCM trajectory, the following control law is chosen:
\begin{equation}
\label{eq:instanteneous_dcm}
r_{zmp}^{*} = \xi^{ref} - b \dot{\xi}^{ref} + K^{\xi}_{p} (\xi - \xi^{ref}) 
+ K^{\xi}_{i} \int{(\xi - \xi^{ref}) \diff t}.
\end{equation}
By applying the control input defined in \eqref{eq:instanteneous_dcm}, the closed loop dynamics writes:
\begin{equation}
\label{eq:instanteneous_cl}
\dot{\tilde{\xi}} = \dot{\xi} - \dot{\xi}^{ref} = \frac{1}{b}(I_2 - K^{\xi}_{p})  \tilde{\xi} - \frac{1}{b} K^{\xi}_{i}  \int{\tilde{\xi} \diff t}.
\end{equation}
If $K^{\xi}_{p}> I_2$ and $K^{\xi}_{i} > 0_2$ the closed loop dynamics is asymptotically stable, and consequentially the error $\tilde{\xi}$ converges asymptotically to zero. 
\par
The main advantage of using the presented controller is the triviality of its implementation. On the other hand, the controller cannot guarantee the feasibility of the gait since the position of the ZMP may exit the support polygon.

\subsubsection{DCM predictive control}
\label{sec:predictive-control}
To guarantee a feasible ZMP, a model predictive controller can be designed.\cite{Krause2012} In the MPC framework, the DCM dynamics \eqref{eq:dcm_dynamics} is used as a prediction model and it is discretized assuming piecewise constant ZMP trajectories with a constant sampling time $T$:
\begin{equation}
\label{eq:dcm_discrete}
\xi_{k+1} = F \xi_k + G r_k^{zmp} = e^{\frac{T}{b}} \xi_k + (1 - e^{\frac{T}{b}}) r_k^{zmp}.
\end{equation}
In order to ensure that the stance foot does not rotate around one of its edges, the desired ZMP must not exit the support polygon.\cite{Vukobratov2004} This is ensured through a set of linear inequality constraints:
\begin{equation}
\label{eq:inequality_constraints}
A_{c_k} r^{zmp}_k \le b_{c_k}.
\end{equation}
\par
The tracking of the desired DCM trajectory and a smooth control signal are obtained with the following cost function:
\begin{equation}
\begin{split}
    H_k & =  \sum\limits_{j=k}^{N + k-1} ({\xi}_j-{\xi}_j^{ref})^\top Q ({\xi}_j-{\xi}_j^{ref}) +  ({r}^{zmp}_j - {r}^{zmp}_{j-1})^\top R ({r}^{zmp}_j - {r}^{zmp}_{j-1}) \\
& + ({\xi}_{k+N}-{\xi}_{k+N}^{ref})^\top Q_N ({\xi}_{k+N}-{\xi}_{k+N}^{ref}),
\end{split}
 \label{eq:mpc_cost}
\end{equation}
where $Q$, $Q_N$ and $R$ are symmetric positive definite matrices and $N$ is the length of the preview window. The cost \eqref{eq:mpc_cost} can be split into three terms. The first and the third are related to the tracking of the desired DCM trajectory while the second one tries to minimize the rate of change of the control input $r^{zmp}$.
Since the cost function is a quadratic positive function and the constraints are linear, the optimal control problem a strictly convex quadratic programming problem (QP).

\subsubsection{ZMP-CoM Controller}
\label{sec:ZMP-CoM-Controller}
Both controllers defined in Section~\ref{sec:instantaneous-control} and \ref{sec:predictive-control} provide a desired ZMP position. In case of kinematic based whole-body QP control layer, another control loop is needed in order to obtain such desired ZMP position. Instead, as explained later in Section~\ref{sec:dynamics_QP}, if the whole-body QP control layer is developed by taking into account the dynamics of the robot, such additional control loop is not necessary. The ZMP tracking problem is tackled by implementing the control law, \cite{Choi2007} i.e.:
\begin{equation}
\label{eq:zmp_controller}
v_\mathcal{C}^* = v_\mathcal{C}^{ref} - K_{zmp}(r_{zmp}^{*} - r_{zmp}) + K_{com} (p^{ref}_\mathcal{C} - p_\mathcal{C}),
\end{equation}
where $K_{com} > b^{-1} I_2$  and $0_2 < K_{zmp} < b^{-1} I_2.$

\subsection{Whole-body QP control layer}
The main objective of the whole-body QP control layer is to ensure the tracking of the desired trajectories by using complete robot models. 
In the following sections, we analyze two kinds of whole-body QP controllers. First, we design a controller based only on the kinematic model. Secondly, we tackle the control problem by means of the dynamic model.

\subsubsection{Kinematics based whole-body QP control layer}
The goal of the kinematics based whole-body QP control layer is to ensure the tracking of the position of the CoM, the feet pose and the torso orientation. The control problem is formulated using the stack of tasks approach. The tracking of the feet and of the CoM trajectories are considered as high priority tasks, while the torso orientation is considered as a low priority task. 
Furthermore, to attempt the stabilization of the zero dynamics of the system,\cite{nava16} a postural condition is added as a low-priority task.
\par
The control objective is achieved by framing the controller as a constrained optimization problem where the low priority tasks are embedded in the cost function and the high priority tasks in the constraints. The cost function is given by:
\begin{equation}
\label{eq:velocity_control_cost}
h(\nu) = \frac{1}{2} \left[ (\omega ^ {des} _ {\mathcal{T}} - J_{\mathcal{T}} \nu)^\top
\Lambda_{\mathcal{T}} (\omega ^ {des} _ {\mathcal{T}} - J_{\mathcal{T}} \nu) + (\dot{s} - \dot{s}^{des} ) ^\top \Lambda (\dot{s} - \dot{s}^{des})\right], 
\end{equation}
where $\Lambda_{\mathcal{T}} > 0$, $\Lambda > 0$.
The tracking of the desired torso orientation is achieved by the first term of \eqref{eq:velocity_control_cost} with the desired torso angular velocity $\omega^{des}_{\mathcal{T}}$. By a particular choice of such desired velocity, it is possible to guarantee almost-global stability and convergence of ${}^\mathcal{I}R _{\mathcal{T}}$ to ${}^\mathcal{I}R _{\mathcal{T}}^{ref}$.\cite{Olfati-Saber:2001:NCU:935467}
The second term of \eqref{eq:velocity_control_cost} is the postural task. It is achieved by specifying a desired joints velocity that depends on the error between the desired and measured joints position 
\begin{equation}
\label{eq:regularization_term}
\dot{s}^{des} = -K_{s} (s - s^{ref}) \quad K_{s} > 0.
\end{equation}
\par
The hard constraints are:
\begin{equation}
\label{eq:velocity_control_jacobians}
J_{\mathcal{C}}(\nu) \nu = v^ {des} _ {\mathcal{C}}, \quad
J_{\mathcal{F}}(\nu) \nu = \mathrm{v}^ {des} _ {\mathcal{F}},
\end{equation}
where $v^{des}_{\mathcal{C}}$ is the linear velocity of the CoM, $\mathrm{v}^{des}_{\mathcal{F}}$ is the desired foot twist. More specifically the foot velocities $\mathrm{v}^{des}_{\mathcal{F}}$ are chosen as:
\begin{equation}
\label{feetVelocitiesStar}
\mathrm{v}^{des}_ {\mathcal{F}} = \prescript{}{}{\mathrm{v}}^{ref}_{\mathcal{F}} -
\begin{bmatrix}
K^p _{x} (p_{\mathcal{F}} - p^{ref}_{\mathcal{F}})
+ K^i _{x }\int{ (p_{\mathcal{F}} - p^{ref}_{\mathcal{F}}) \diff t}\\
K _{\omega} \left[\text{sk}(\prescript{\mathcal{I}}{}{R}_{\mathcal{F}} \prescript{\mathcal{I}}{}{R} _{\mathcal{F}} ^{ref^\top} )\right]^{\vee}
\end{bmatrix}.
\end{equation}
Here the gain matrices are positive definite. The desired position $p^{ref}_{\mathcal{F}}$, orientation $\prescript{\mathcal{I}}{}{R} _{\mathcal{F}} ^{ref}$ and the velocities $\prescript{\mathcal{I}}{}{\mathrm{v}}^{ref}_{\mathcal{F}}$ are the output of the trajectory optimization layer.
\par
The desired CoM velocity $v^ {des} _{\mathcal{C}}$ is chosen as:
\begin{equation}
v^{ref}_{\mathcal{C}} = v_{\mathcal{C}}^* - K^p _{\mathcal{C}}(p_{\mathcal{C}} - p_{\mathcal{C}}^*) - K^i _{\mathcal{C}}\int{ p_{\mathcal{C}} - p_{\mathcal{C}}^*  \diff t},
\end{equation}
where the gain matrices are positive definite, $\dot{x}^*$ is the output of the ZMP-CoM \eqref{eq:zmp_controller} controller and $x^*$ is the integrated signal. 
\par
Finally, an inequality constraint is added to limit the maximum joint velocities.
\begin{equation}
\label{eq:ik_jacobian_limits}
\dot{s}^- \le \dot{s} \le \dot{s}^+.
\end{equation}
\par
Since~\eqref{eq:velocity_control_cost} and \eqref{eq:velocity_control_jacobians} are respectively quadratic and linear with respect to the decision variable $\nu$, the control problem can be converted into a QP problem.\cite{8625025}

The decision variable $\nu$ contains the desired joint velocities. This desired quantity can be used directly as a reference to a joint velocity controller, if available. Otherwise, they can be integrated and fed to a low-level joints position controller.

\subsubsection{Dynamics-based whole-body QP Control Layer} \label{sec:dynamics_QP}
The Dynamics-based whole-body QP control layer uses the dynamic model of the system to ensure the tracking of the desired trajectories. The control problem is formulated using the stack of tasks approach. The high priority tasks are the tracking of the desired feet and the ZMP trajectories, while the torso orientation and the postural task are still considered as a low priority.
\par
The control objective is achieved by designing the controller as a constrained quadratic optimization problem whose conditional variables are 
$
u = \begin{bmatrix} \dot{\nu}^\top & \tau^\top & f^\top \end{bmatrix} ^ \top
$.
\par
The following cost function holds:
\begin{subequations}
\begin{align}
    h(u)= \frac{1}{2} & \left[  (\dot{\omega}_{\mathcal{T}} ^ {des} - \dot{\omega}_{\mathcal{T}}) ^ \top \Lambda_{\mathcal{T}} (\dot{\omega}_{\mathcal{T}} ^{des} - \dot{\omega}_{\mathcal{T}}) \right. \label{eq:torque_control_cost_torso}\\
    &  + (\ddot{s}\,^{des} - \ddot{s}) ^ \top \Lambda _ s (\ddot{s}\,^{des} - \ddot{s}) \label{eq:torque_control_cost_regularization}\\
    &  + \tau ^ \top \Lambda \tau \label{eq:torque_control_cost_torque}  \\
    &  + \left. f ^ \top \Lambda _ f f \right] \label{eq:torque_control_cost_force} 
\end{align}
\end{subequations}
where the term \eqref{eq:torque_control_cost_torso} is in charge of stabilizing the desired torso orientation:
\begin{equation}
\label{eq:rotational_pid_acceleration}
\begin{split}
\dot{\omega}^{des}_\mathcal{T} &= \dot{\omega}^{ref} - c_0 \left(\hat{\omega} \prescript{\mathcal{I}}{}{R}_{\mathcal{T}} \prescript{\mathcal{I}}{}{R}_{\mathcal{T}} ^{ref ^ \top} - \prescript{\mathcal{I}}{}{R}_{\mathcal{T}} \prescript{\mathcal{I}}{}{R}_{\mathcal{T}} ^{ref^{\top}} \hat{\omega}^{ref}\right)^{\vee} \\
&- c_1 \left(\omega - \omega^{ref}\right) - c_2 \left(\prescript{\mathcal{I}}{}{R}_{\mathcal{T}} \prescript{\mathcal{I}}{}{R}_{\mathcal{T}} ^{{ref}^{\top}}\right) ^\vee.
\end{split}
\end{equation}
Here $c_0$, $c_1$ and $c_2$ are positive numbers.
The postural task \eqref{eq:torque_control_cost_regularization} is achieved by asking for a desired joint acceleration that depends on the error between the desired and measured joint values:
\begin{equation}
    \ddot{s} ^{des} = \ddot{s} ^ * - K ^ d _s (\dot{s} - \dot{s} ^{ref}) - K ^ p _s ( s  - s^ {ref}).
\end{equation}
Lastly~\eqref{eq:torque_control_cost_torque} and~\eqref{eq:torque_control_cost_force} are used to minimize the joint torques and forces required by the controller.
\par
The hard constraints are the tracking of the feet, the ZMP and the control of the CoM height. Since the conditional variables contain both the robot acceleration, the joint torques and the contact wrenches, the robot dynamics is also added as equality constraint. Finally, the feasibility of the desired contact wrenches $f$ is guaranteed via another set of inequalities. More specifically $f$ has to belong to the associated friction cone, while the position of the local CoP is constrained within the support polygon. 
The feasibility of the contact wrenches is represented by a linear inequality constrain of the form: $B_f f \le 0$.
\par
Concerning the tracking of the feet and the CoM height, we have:
\begin{equation}
J_{\circ} \dot{\nu} = \dot{\mathrm{v}}^{des}_{\circ} - \dot{J}_{\circ} \nu \quad \quad  \circ = \{\mathcal{F}, \mathcal{C}\}.
\end{equation}
When the foot is in contact, the desired acceleration $\dot{\mathrm{v}}^{des}_{\mathcal{F}}$ is zero. During the swing phase, the angular part of $\dot{\mathrm{v}}^{des}_{\mathcal{F}}$ is given by \eqref{eq:rotational_pid_acceleration} where the subscript $\mathcal{T}$ is substitute with $\mathcal{F}$, while the linear part $\dot{v}^{des}$ is equal to:
\begin{equation}
    \label{eq:torque_feet_linear_pid}
    \dot{v}^{des}_{\mathcal{F}} =  \dot{v}\,^{ref}_{\mathcal{F}} - K^d _{x _{f}} (v_{\mathcal{F}} - v^{ref}_{\mathcal{F}}) - K^p _{x _{f}} (p_{\mathcal{F}} - p^{ref}_{\mathcal{F}}).
\end{equation}
Here the gains are again positive definite matrices. The control law in \eqref{eq:torque_feet_linear_pid} holds also for the CoM height by using the corresponding quantities.
\par
The tracking of the desired global ZMP, evaluated either by the controller in Sec~\ref{sec:instantaneous-control} or in \ref{sec:predictive-control}, is ensured by an equality constraint. In case of single support phase, it is given by:
\begin{equation}
\begin{bmatrix}
0 & 0 & r ^ {ref} _{{zmp}_ x} - p_{\mathcal{F}_x} & 0 & 1 & 0\\
0 & 0 & r ^ {ref} _{{zmp}_ y} - p_{\mathcal{F}_y} & -1 & 0 & 0
\end{bmatrix} f = A_{ss} f = 0,
\end{equation}
where the subscripts $x$ and $y$ indicates the $x$ and $y$ coordinates of the vectors. 
\par
By rearranging \eqref{eq:robot_dynamic}, the robot dynamics can be treated as an equality constraint of the following form:
\begin{equation}
    C(q, \nu)\nu + G(q) = 
    \begin{bmatrix}
    - M(q) & B & J^\top_{c}(q)
    \end{bmatrix} u = H(q) u,
\end{equation}
where $J_{c}(q)$ is the matrix contained the contact wrenches Jacobians.
\par 
Finally, an inequality constraint is added to limit the maximum joint torques
\begin{equation}
\label{eq:torque_limits}
\tau^- \le \tau \le \tau^+.
\end{equation}
The controller presented above can be represented as a QP problem and solved via off-the-shelf solvers.

\subsubsection{Floating Base estimation}
The floating base pose and velocity are crucial for torque-controlled walking and they are not directly measurable on the real robot. Thus, they need to be estimated using a simple base estimation framework. In order to estimate these quantities, we assume that at least one link of the robot is rigidly attached to the environment at every instant of time. 

The floating base pose, described by the frame transformation matrix $\prescript{\mathcal{I}}{}{H}_\mathcal{A}$, is computed fusing legged odometry and contact switching information. 
The pose $\prescript{\mathcal{I}}{}{H}_\mathcal{F}$ of the fixed link $\mathcal{F}$ w.r.t. the inertial frame $\mathcal{I}$ at the first time-instant, $t = 0$, is known. At this instant, the base pose can simply be computed as,
\begin{equation}
    \label{eq:legged-odom}
    \prescript{\mathcal{I}}{}{H}_\mathcal{B} = \prescript{\mathcal{I}}{}{H}_\mathcal{F}\prescript{\mathcal{F}}{}{H}_\mathcal{B}(s),
\end{equation}
where, the relative transform $\prescript{\mathcal{F}}{}{H}_\mathcal{B}$ between the fixed link $\mathcal{F}$  and the base $\mathcal{B}$ can be obtained through forward kinematics using the encoder measurements. The current transform $\prescript{\mathcal{I}}{}{H}_\mathcal{F}$ is stored and assumed to be fixed until a contact switch is triggered. As soon as a contact switch occurs, the fixed link is changed to the new link in rigid contact with the environment and the base transform is updated as:
\begin{equation}
    \label{eq:base-pose}
    \prescript{\mathcal{I}}{}{H}_\mathcal{B} = \prescript{\mathcal{I}}{}{H}_\mathcal{F_{\text{old}}}\prescript{\mathcal{F_{\text{old}}}}{}{H}_\mathcal{F_{\text{new}}}(s)\prescript{\mathcal{F_\text{new}}}{}{H}_\mathcal{B}(s),
\end{equation}

The information about the contact switching is obtained through a Schmitt Trigger thresholding on the contact normal forces. 
The contact normal forces are, in turn, obtained from the end-effector wrenches estimated by the external wrench estimation method described in.\cite{Nori2015}
\par
The floating base velocity, $\prescript{}{}{\mathrm{v}}_\mathcal{B}$, is computed considering the constraint that, when a link is rigidly attached to the environment, the velocity of such link is zero. As a consequence, the floating system velocity $\nu$ can be computed through the free-floating Jacobian of the contact link, $J_{\mathcal{F}}$:
\begin{equation}
    \prescript{}{}{\mathrm{v}}_\mathcal{F} = J_{\mathcal{F}} \nu = 0.
\end{equation}

By expressing the Jacobian $J_{\mathcal{F}}$, in terms of base velocity $J_{\mathcal{F}_b}$ and joint velocities $J_{\mathcal{F}_s}$, the floating base velocity can be computed as:
\begin{equation}
    \prescript{}{}{\mathrm{v}}_\mathcal{B} = - J_{\mathcal{F}_{b}}^{-1} J_{\mathcal{F}_{s}} \dot{s}.
\end{equation}

It should be noted that $J_{\mathcal{F}_{b}}$ is a square matrix and it is always invertible. %

\section{Experimental Results}
\label{sec:experimental_results}
In this section, we present experiments obtained from several implementations of the control architecture shown in Fig.~\ref{fig:controller_architecture}. The experimental activities are carried out with the iCub,\cite{Metta2010} a $\SI{104}{\centi \meter}$ tall humanoid robot, with a foot length and width of  $\SI{19}{\centi \meter}$ and $\SI{9}{\centi \meter}$, respectively.
\begin{table}[b]
\centering
\tbl{Maximum straight walking velocities achieved using different implementa-\\tions of the kinematics based control architecture. \label{tab:max_velocity}}
{\begin{tabular}{@{}cc|c@{}} \toprule
Simplified Model Control & Whole-Body QP Control & Max Straight Velocity (m/s) \\
\colrule
Predictive  & Velocity  &  0.1563\\
Predictive  & Position  & 0.1645\\
Instantaneous  & Velocity  &  0.1809\\
Instantaneous  & Position  & 0.3372 \\
\botrule
\end{tabular}}
\end{table}
The optimization problems are solved by using the OSQP library.\cite{osqp} The time horizon of the predictive control described in \ref{sec:predictive-control} is $\SI{2}{\second}$.

Comparing different control architectures, however, is a far cry from being an easy task. Thus, we decided to follow a similar approach presented by Torricelli et al.\cite{Torricelli}
In all the experiments the humanoid robot walks on a horizontal ground at a constant speed. 
In the following sections, we benchmark the different implementations of the controller architecture focusing on two main aspects: tracking and energy consumption performances. 

In our previous work, \cite{8625025} the benchmarking versus the walking velocity was performed by considering the desired velocity set in the Trajectory Optimization layer. Since the Trajectory Optimization layer computes the desired trajectories solving an optimization problem, the actual planned velocity may be different from the velocity set in the layer. Although the measured robot CoM velocity tends to confirm that the data in \cite{8625025} are consistent, the CoM velocity is highly noisy and its significance is still an open point. For this reason, in this paper, we define the walking velocity as the ratio between the step length and the measured step duration. 

\subsection{Tracking Perfomances}
\subsubsection{Kinematics-based Walking Architecture}
\label{sec:kinematics-based_walking_architecture}
Table~\ref{tab:max_velocity} summarizes the maximum velocities achieved using the different implementations of the kinematics based control architecture. In particular, the labels \emph{instantaneous} and \emph{predictive} mean that the associated layer generates its outputs considering inputs and references either at the single time $t$ or for a time window, respectively. The labels, \emph{velocity} and \emph{position} control, instead, mean that the layer outputs are either desired joint velocities or position, respectively.

To compare the kinematics-based controller architectures, we decide to perform two main experiments. In the former, the walking velocity is the one achievable by all the kinematics-based architectures. While, in the latter, the robot walks at the maximum velocity achieved with a specific architecture only -- see Table~\ref{tab:max_velocity}. Namely:
\begin{itemize}
    \item \textbf{experiment 1} the forward robot speed is $\SI{0.1563}{\meter \per \second}$;
    \item \textbf{experiment 2} the forward robot speed is $\SI{0.3372}{\meter \per \second}$.
\end{itemize}
\begin{figure}[t]
    \centering
    \begin{myframe}{Instantaneous + Position Control}
    \begin{subfigure}[b]{0.327\textwidth}
        \centering
        \includegraphics[width=\textwidth]{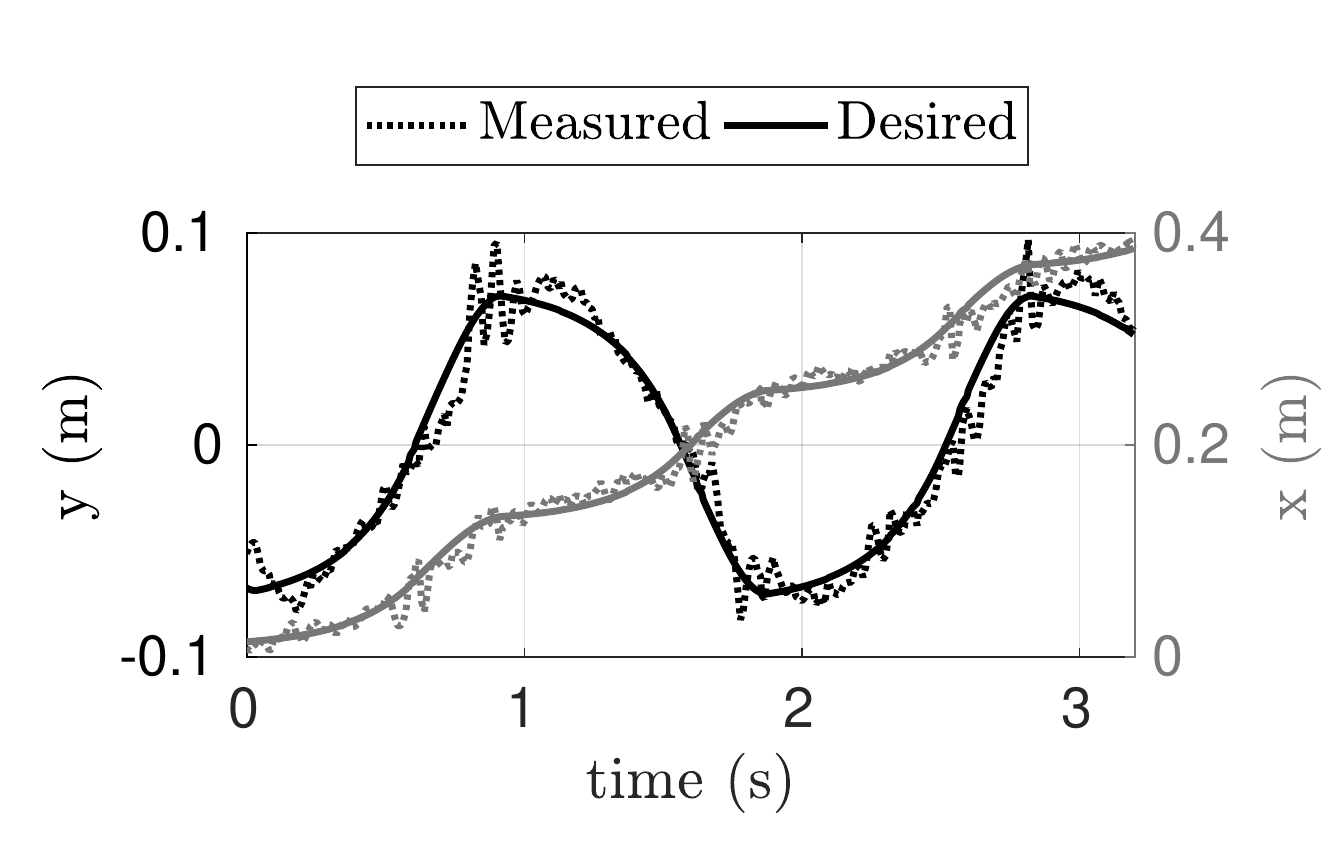}
        \caption{DCM}
        \label{fig:inst_pos-min_vel-dcm}
    \end{subfigure}
    \hfill
    \begin{subfigure}[b]{0.327\textwidth}
        \centering
        \includegraphics[width=\textwidth]{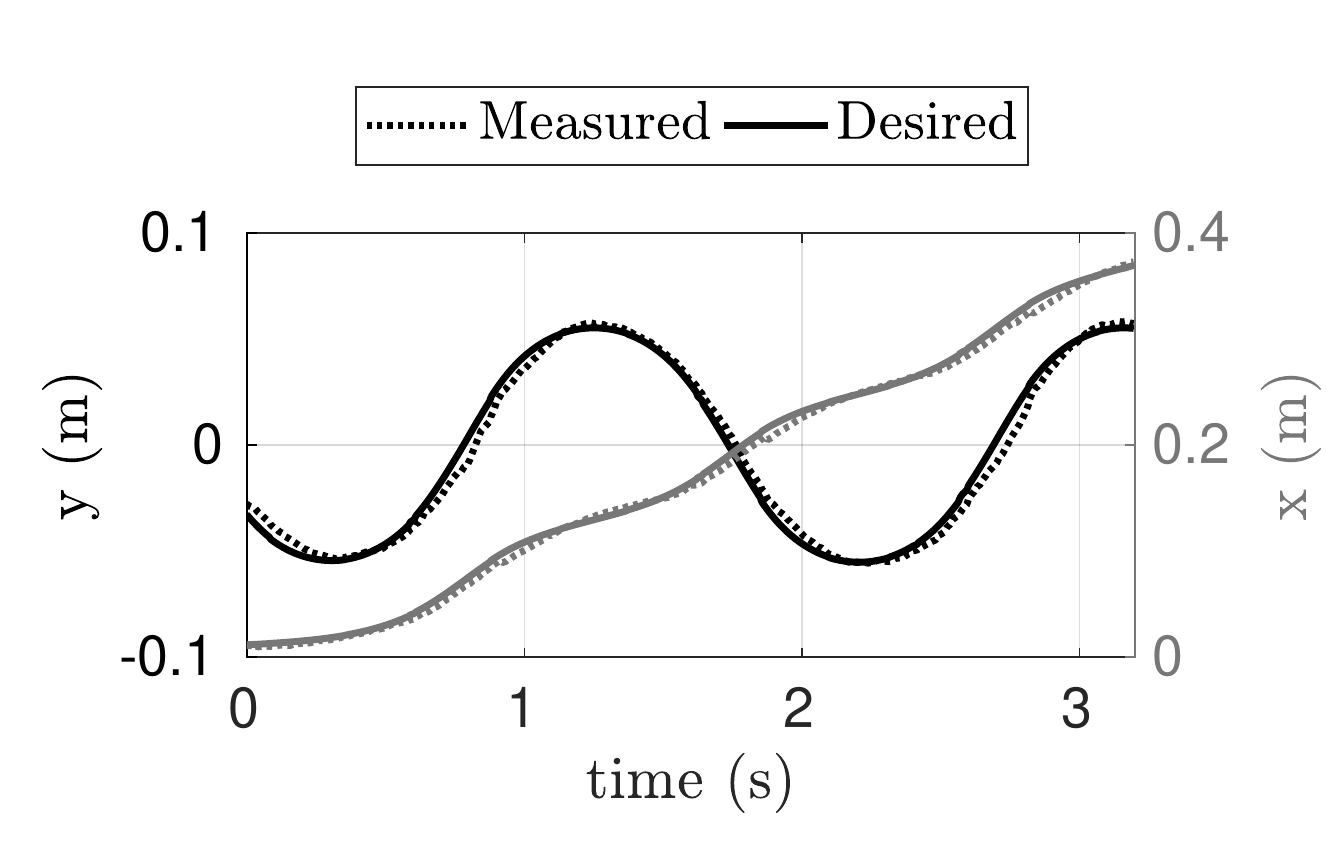}
        \caption{CoM}
        \label{fig:inst_pos-min_vel-com}
    \end{subfigure}
    \hfill
    \begin{subfigure}[b]{0.327\textwidth}
        \centering
        \includegraphics[width=\textwidth]{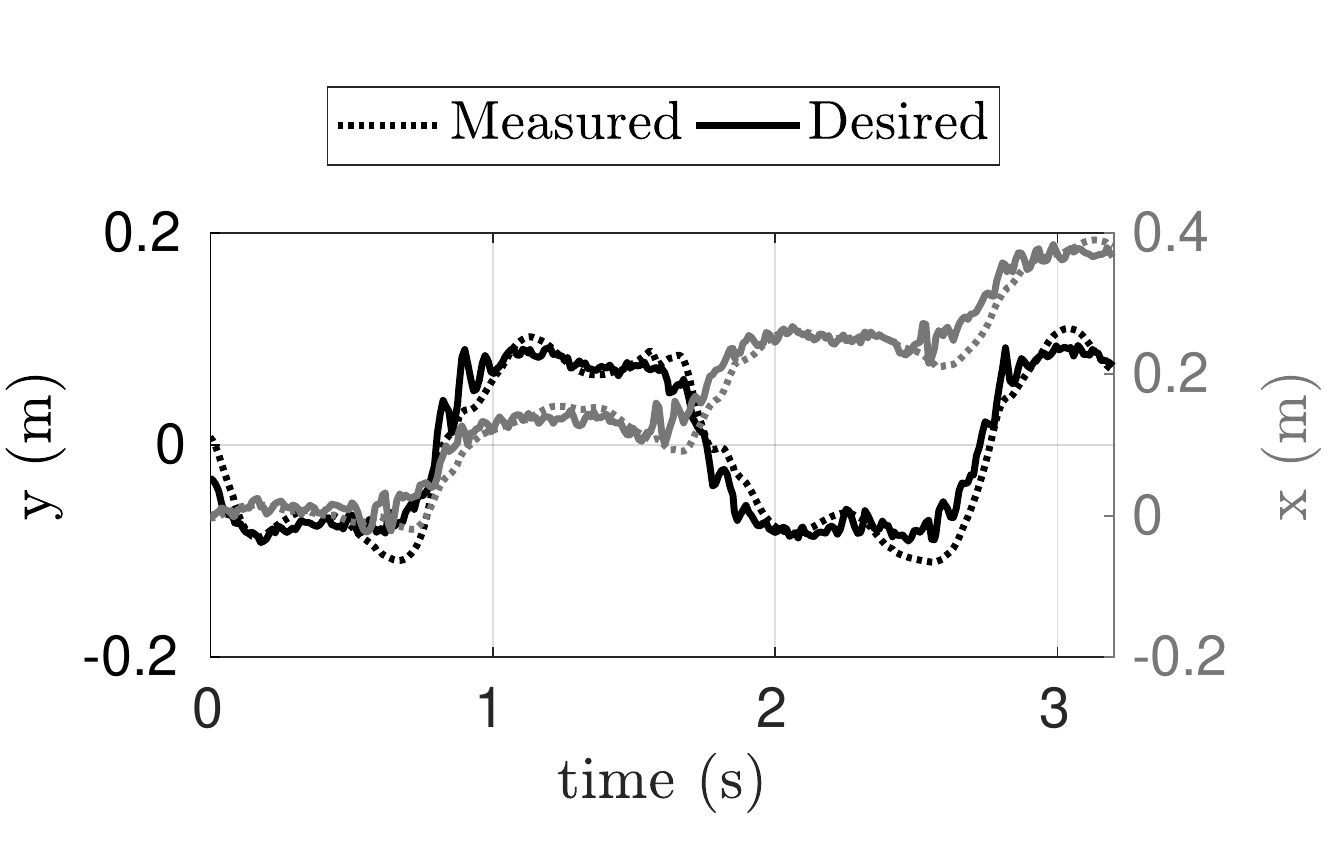}
        \caption{ZMP}
        \label{fig:inst_pos-min_vel-zmp}
    \end{subfigure}
    \end{myframe}
  \hfill
    \begin{myframe}{Predictive + Position Control}
    \begin{subfigure}[b]{0.327\textwidth}
        \centering
        \includegraphics[width=\textwidth]{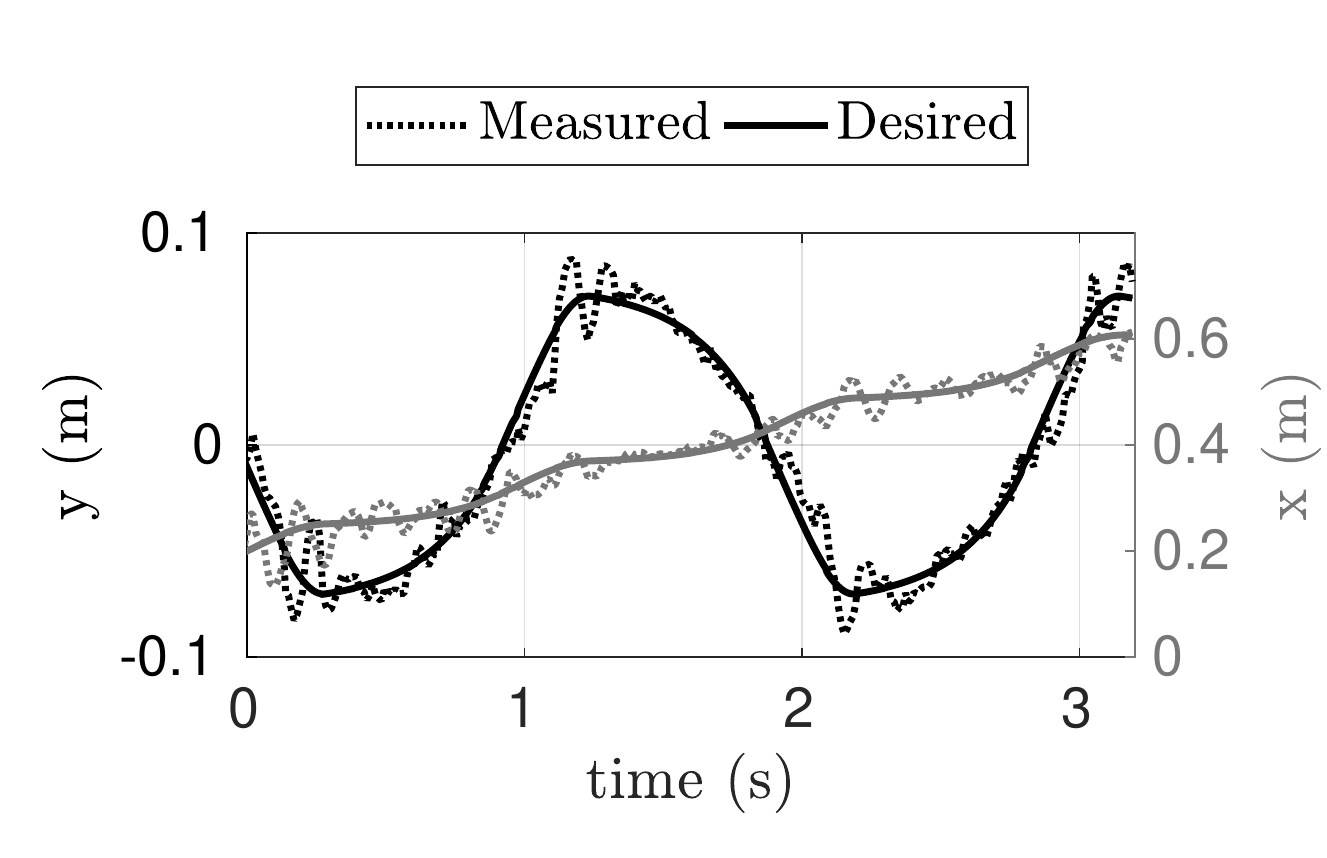}
        \caption{DCM}
        \label{fig:mpc_pos-min_vel-dcm}
    \end{subfigure}
    \hfill
    \begin{subfigure}[b]{0.327\textwidth}
        \centering
        \includegraphics[width=\textwidth]{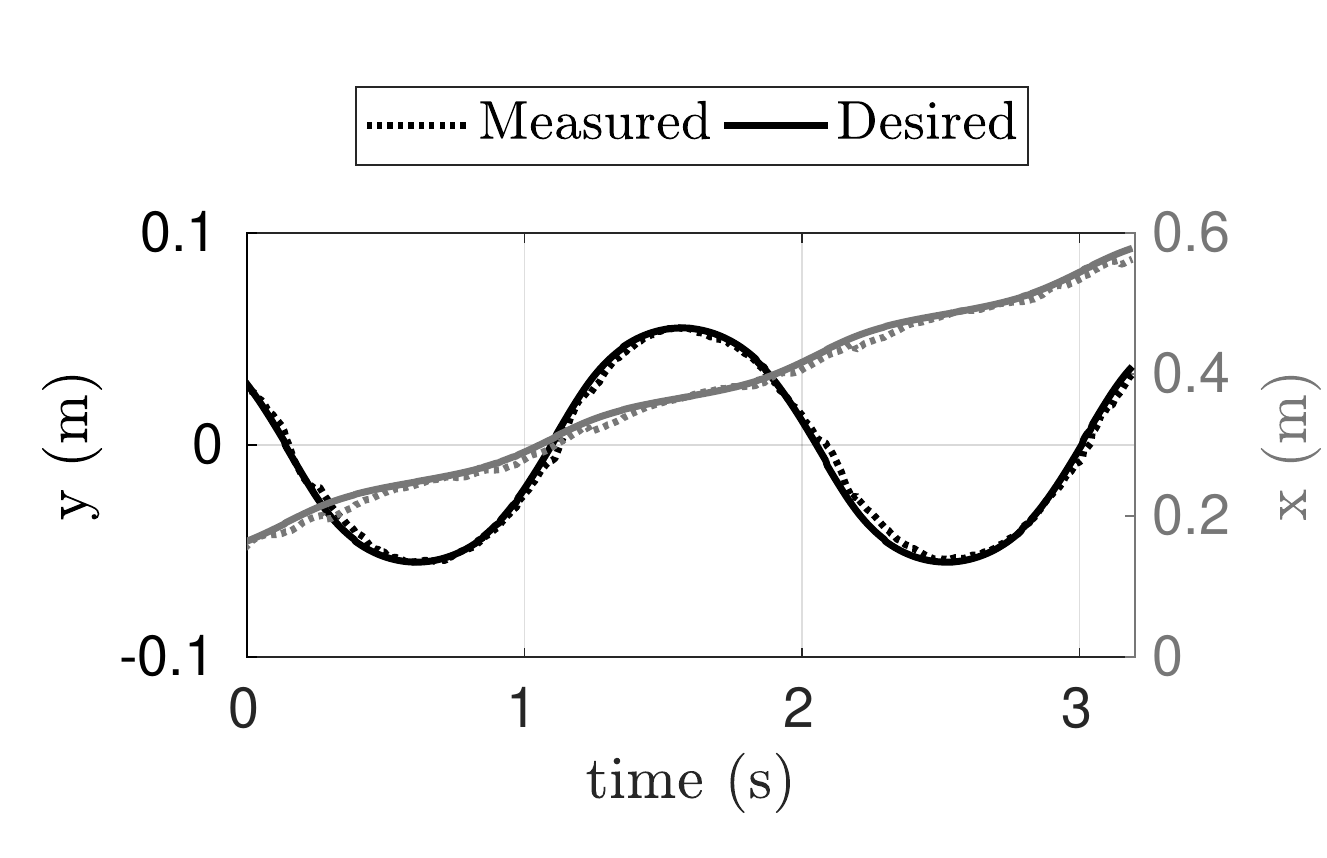}
        \caption{CoM}
        \label{fig:mpc_pos-min_vel-com}
    \end{subfigure}
    \hfill
         \begin{subfigure}[b]{0.327\textwidth}
        \centering
        \includegraphics[width=\textwidth]{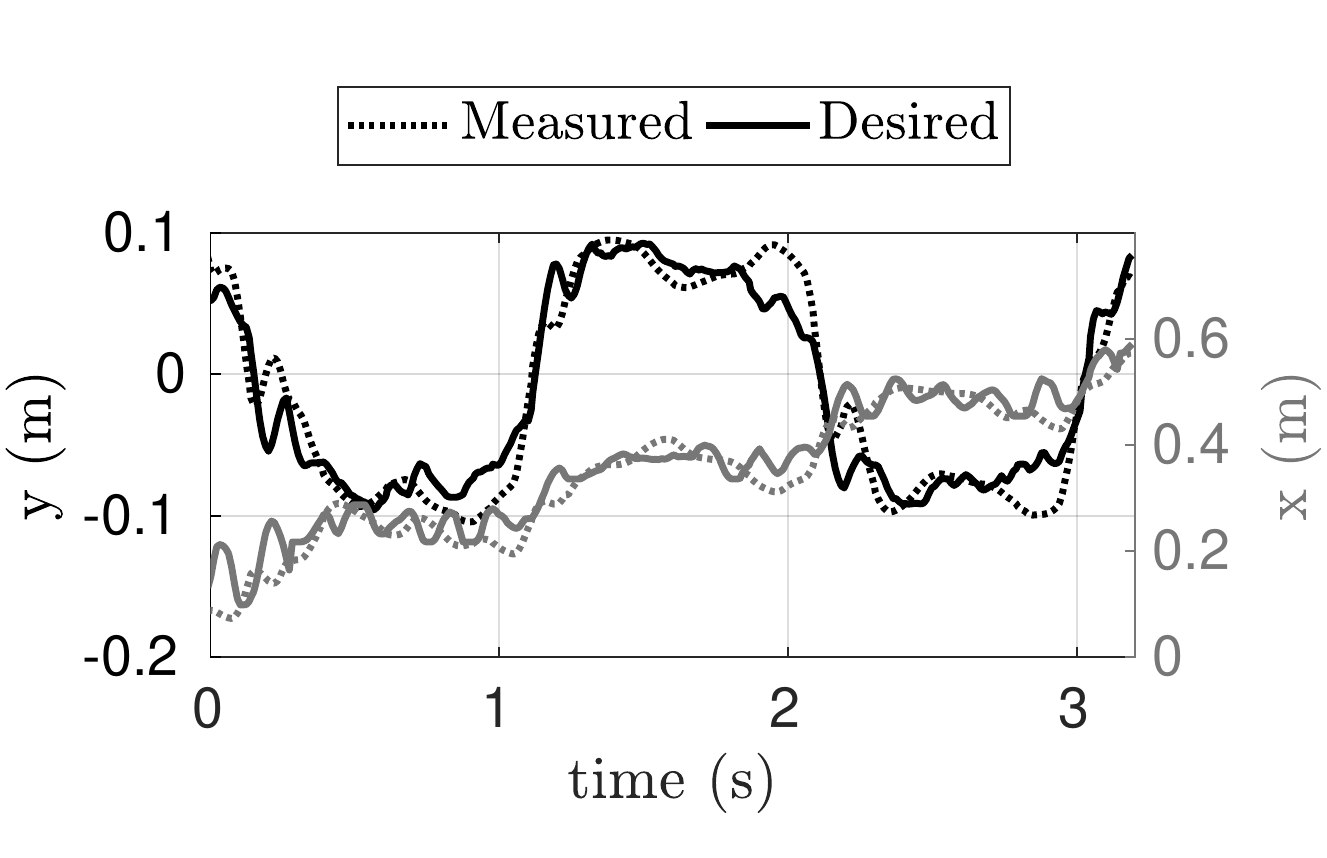}
        \caption{ZMP}
        \label{fig:mpc_pos-min_vel-zmp}
    \end{subfigure}
    \end{myframe}
    \caption{Tracking of the DCM (a), CoM (b) and ZMP (c) using the instantaneous controller with the whole-body controller as position control. Tracking of the  DCM (d), CoM (e) and ZMP (f) using the MPC and the whole-body controller as position control. Forward velocity:  $\SI{0.1563}{\meter \per \second}$.}
\end{figure}
\begin{figure}[t]
    \begin{myframe}{Instantaneous + Position Control}
        \begin{subfigure}[b]{0.327\textwidth}
        \centering
        \includegraphics[width=\textwidth]{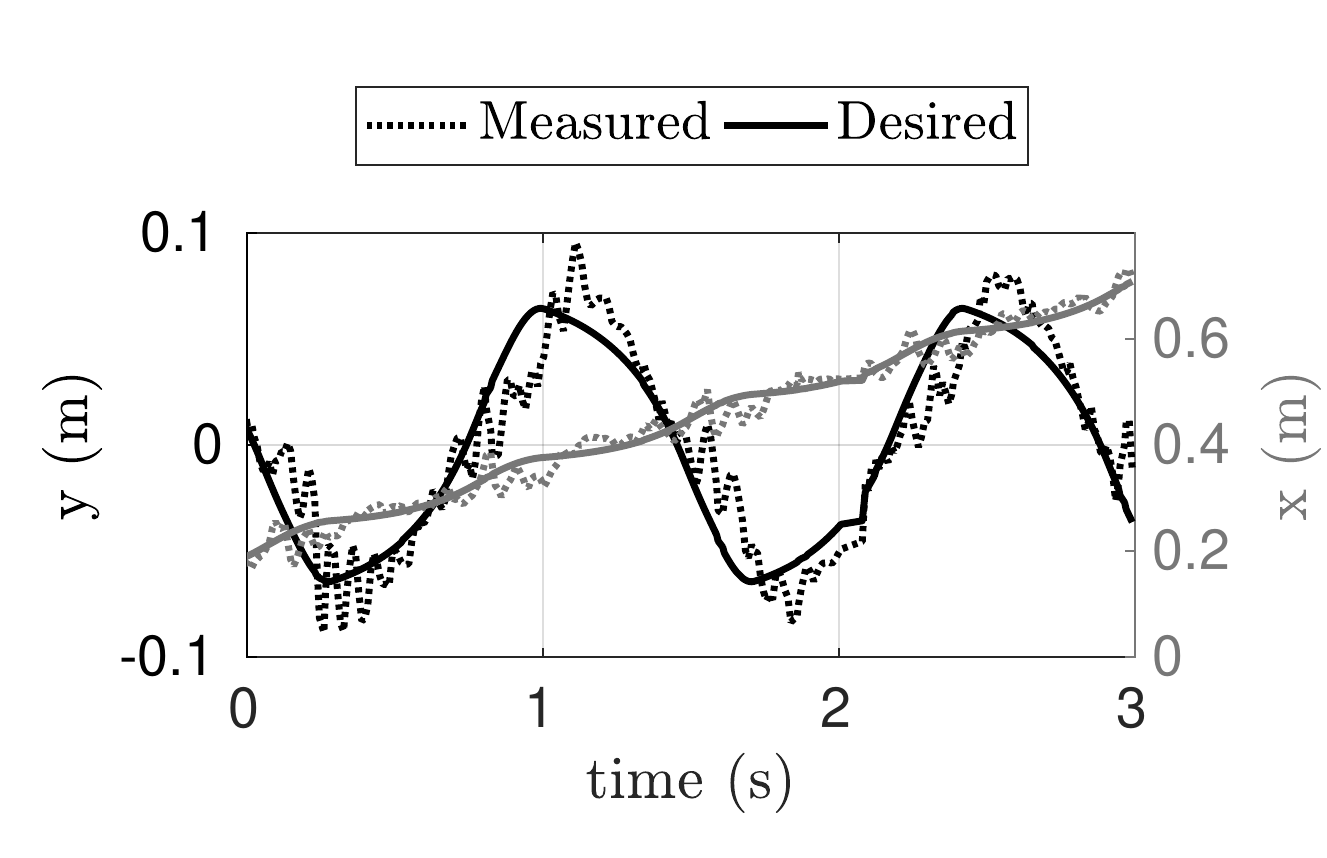}
        \caption{DCM}
        \label{fig:inst_pos-max_vel-dcm}
    \end{subfigure}
    \hfill
     \begin{subfigure}[b]{0.327\textwidth}
        \centering
        \includegraphics[width=\textwidth]{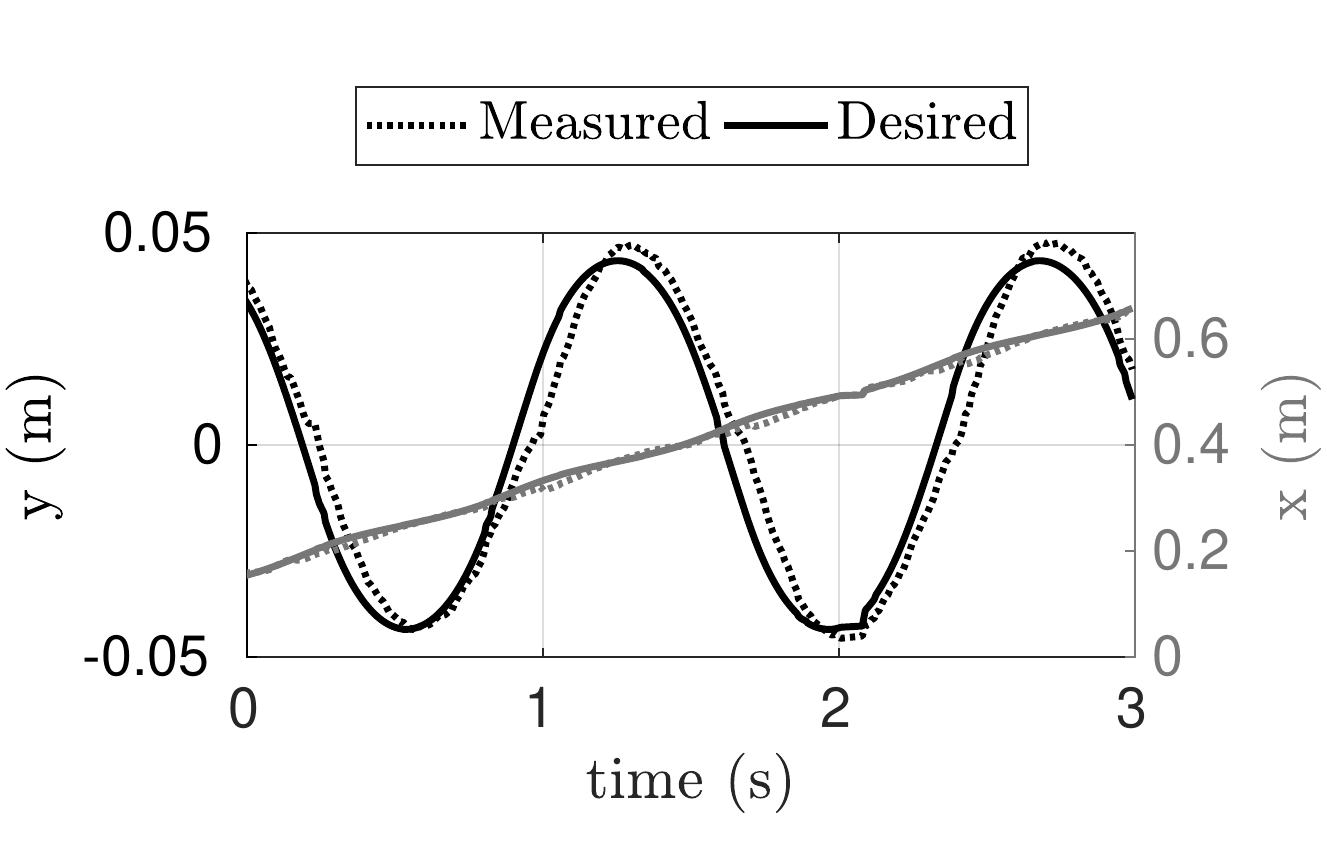}
        \caption{CoM}
        \label{fig:inst_pos-max_vel-com}
    \end{subfigure}
    \hfill
    \begin{subfigure}[b]{0.327\textwidth}
        \centering
        \includegraphics[width=\textwidth]{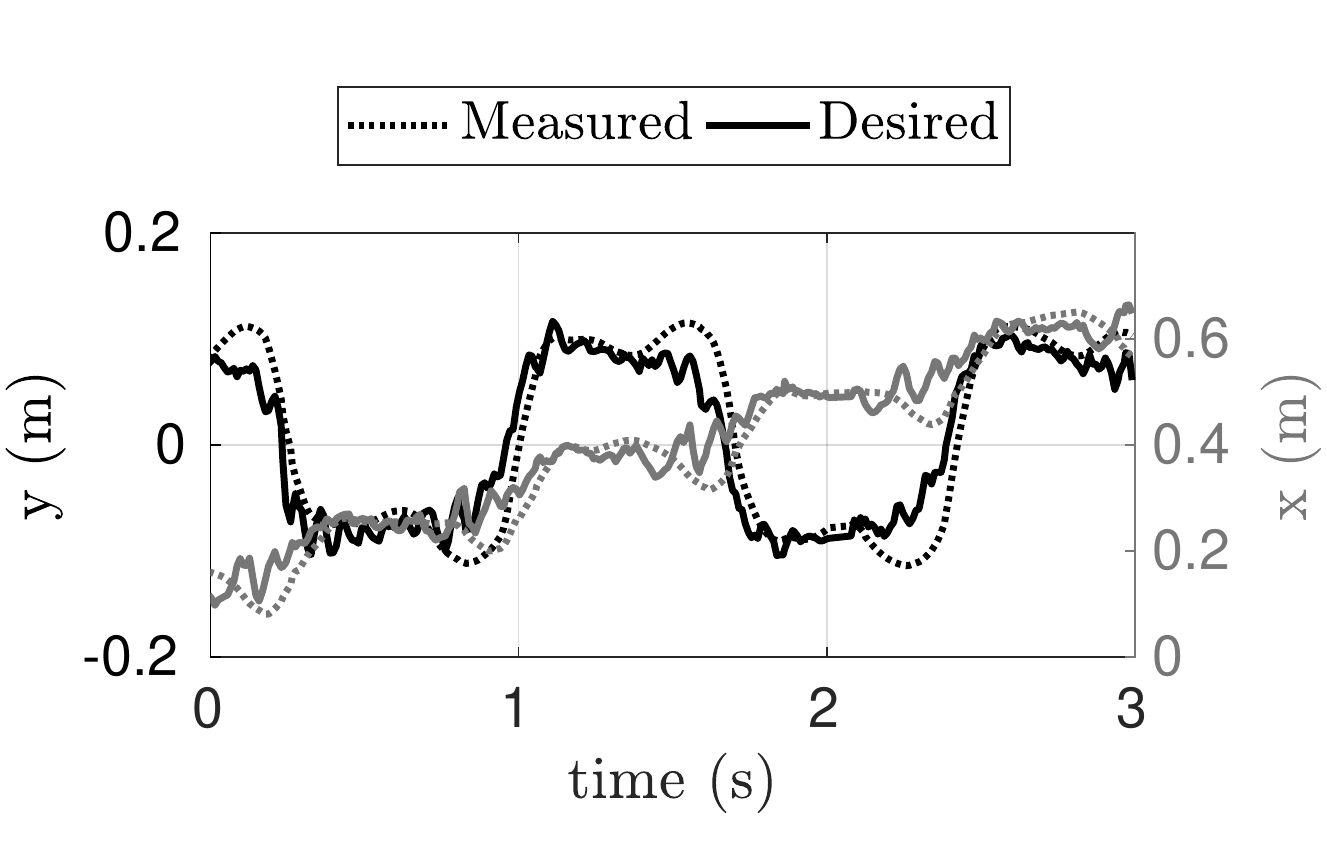}
        \caption{ZMP}
        \label{fig:inst_pos-max_vel-zmp}
    \end{subfigure}
    \end{myframe}
    \hfill
    \begin{myframe}{Predictive + Position Control}
    \begin{subfigure}[b]{0.327\textwidth}
        \centering
        \includegraphics[width=\textwidth]{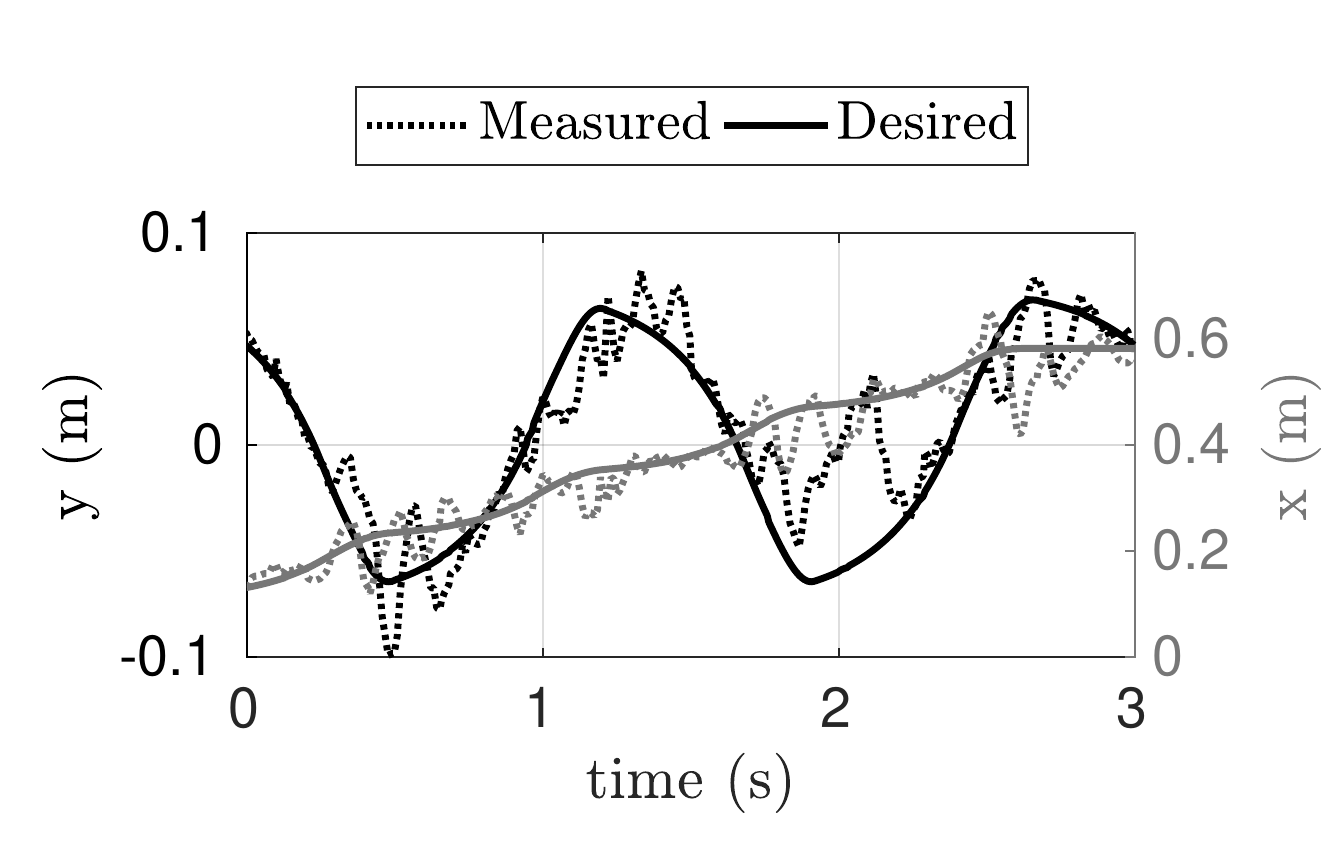}
        \caption{DCM}
        \label{fig:mpc_pos-max_vel-dcm}
    \end{subfigure}
    \hfill
     \begin{subfigure}[b]{0.327\textwidth}
        \centering
        \includegraphics[width=\textwidth]{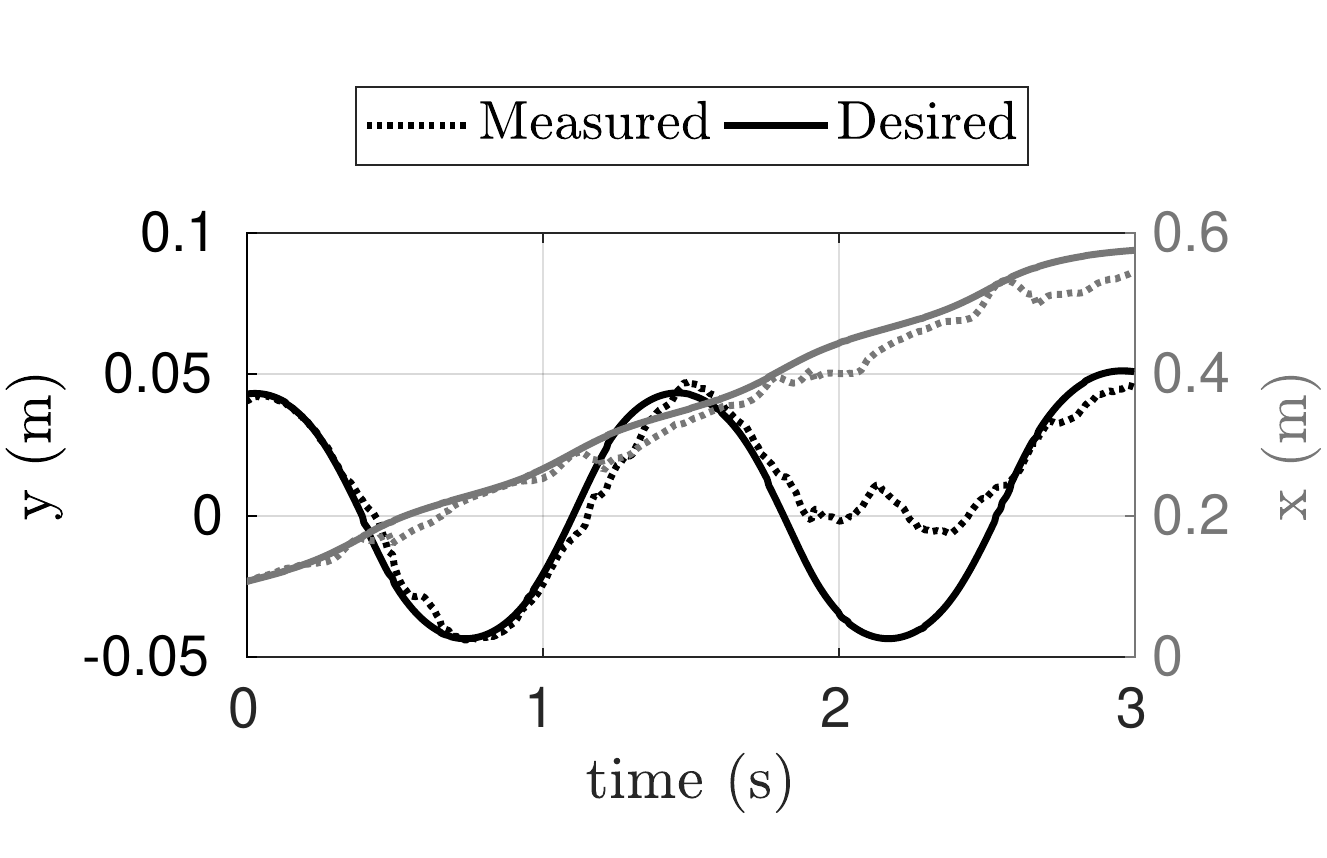}
        \caption{CoM}
        \label{fig:mpc_pos-max_vel-com}
    \end{subfigure}
    \hfill
    \begin{subfigure}[b]{0.327\textwidth}
        \centering
        \includegraphics[width=\textwidth]{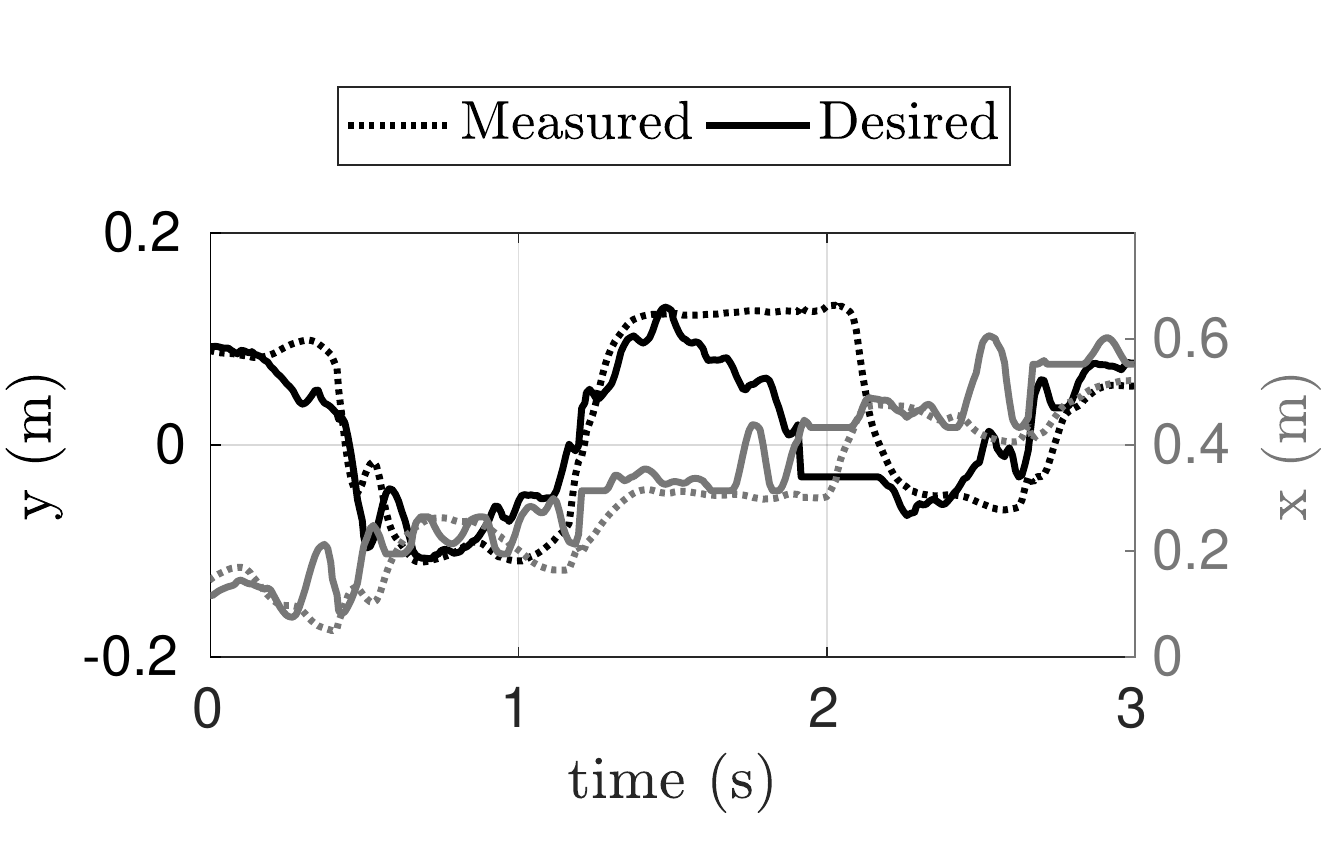}
        \caption{ZMP}
        \label{fig:mpc_pos-max_vel-zmp}
    \end{subfigure}
    \end{myframe}
    \caption{Tracking of the DCM (a), CoM (b) and ZMP (c) with the instantaneous and whole-body QP control as position. Tracking of the  DCM (d), CoM (e) and ZMP (f) with the predictive and whole-body QP control as position control. When the predictive controller was used, at $t\approx \SI{2}{\second}$, the robot falls.  Forward velocity: $\SI{0.3372}{\meter \per \second}$.}
\end{figure}
\begin{figure}[t]
    \centering
    \begin{subfigure}[b]{\textwidth}
        \begin{myframe}{Instantaneous + Position Control}
            \begin{subfigure}[t]{0.49\columnwidth}
            \centering
            \includegraphics[width=\textwidth]{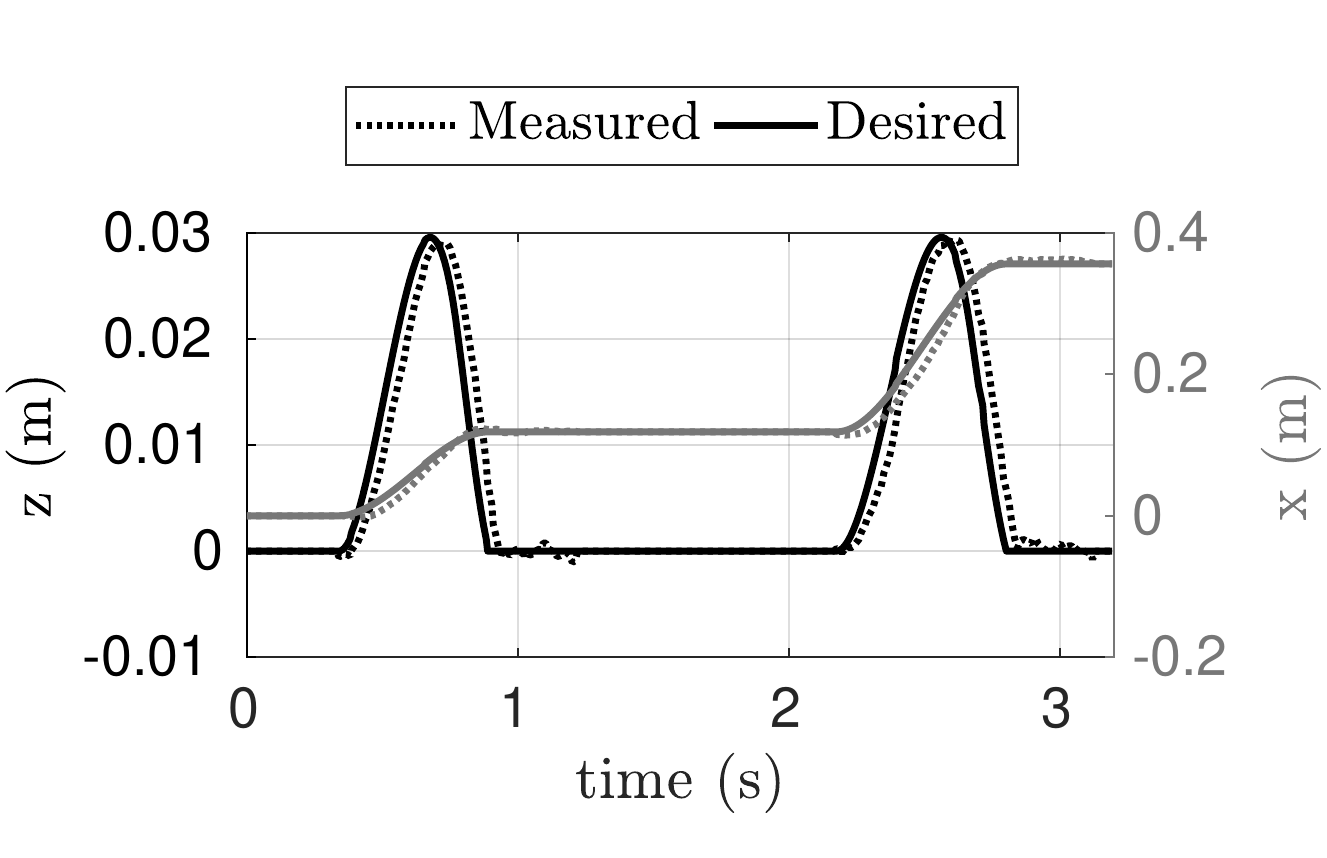}
            \caption{$\SI{0.1563}{\meter \per \second}$}
            \label{fig:inst_pos-min_vel-lf}
            \end{subfigure}
            \begin{subfigure}[t]{0.49\columnwidth}
            \centering
            \includegraphics[width=\textwidth]{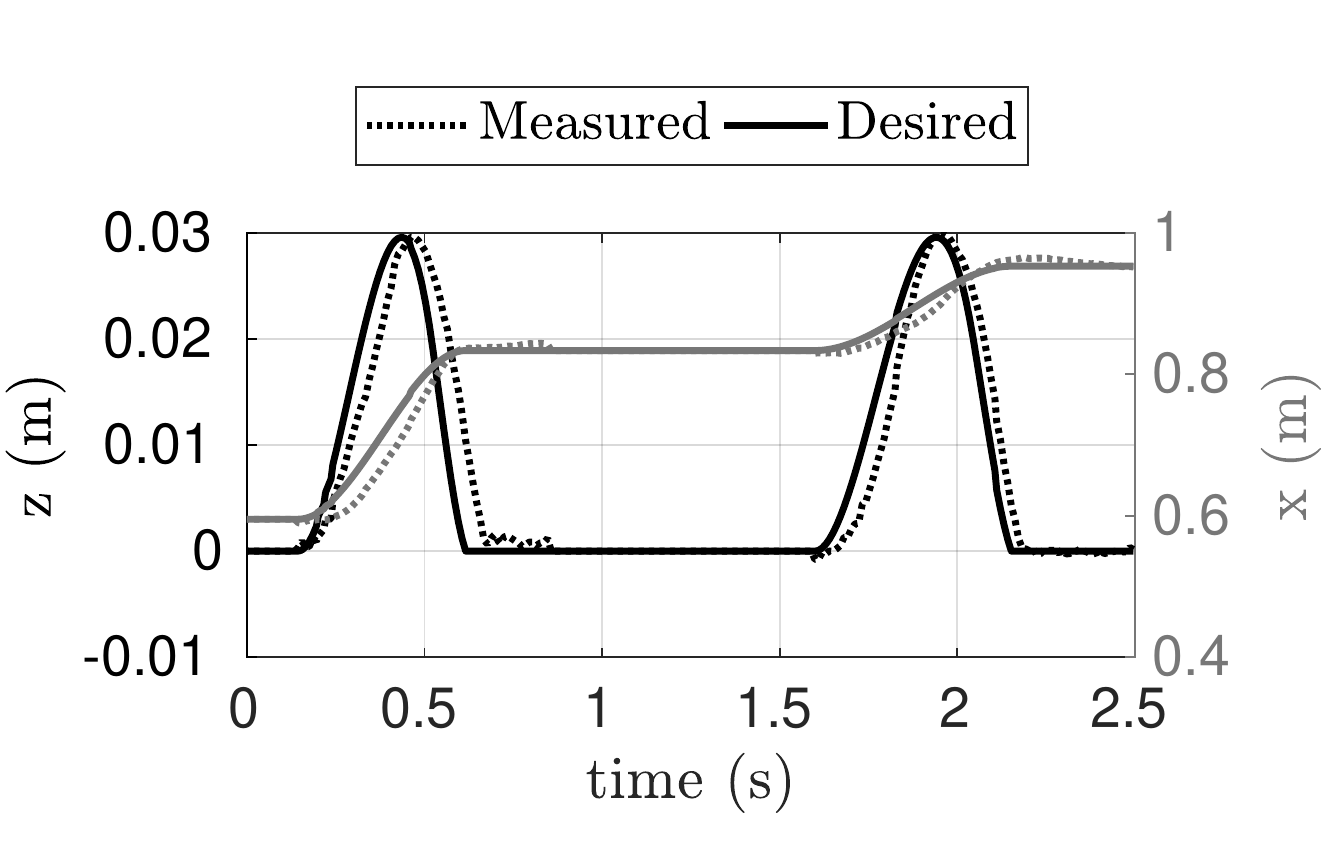}
            \caption{$\SI{0.3372}{\meter \per \second}$}
            \label{fig:inst_pos-max_vel-lf}
            \end{subfigure}
        \end{myframe}
    \end{subfigure}
    \begin{subfigure}[b]{\textwidth}
    \begin{myframe}{Instantaneous + Velocity Control}
        \begin{subfigure}[t]{0.49\columnwidth}
        \centering
        \includegraphics[width=\textwidth]{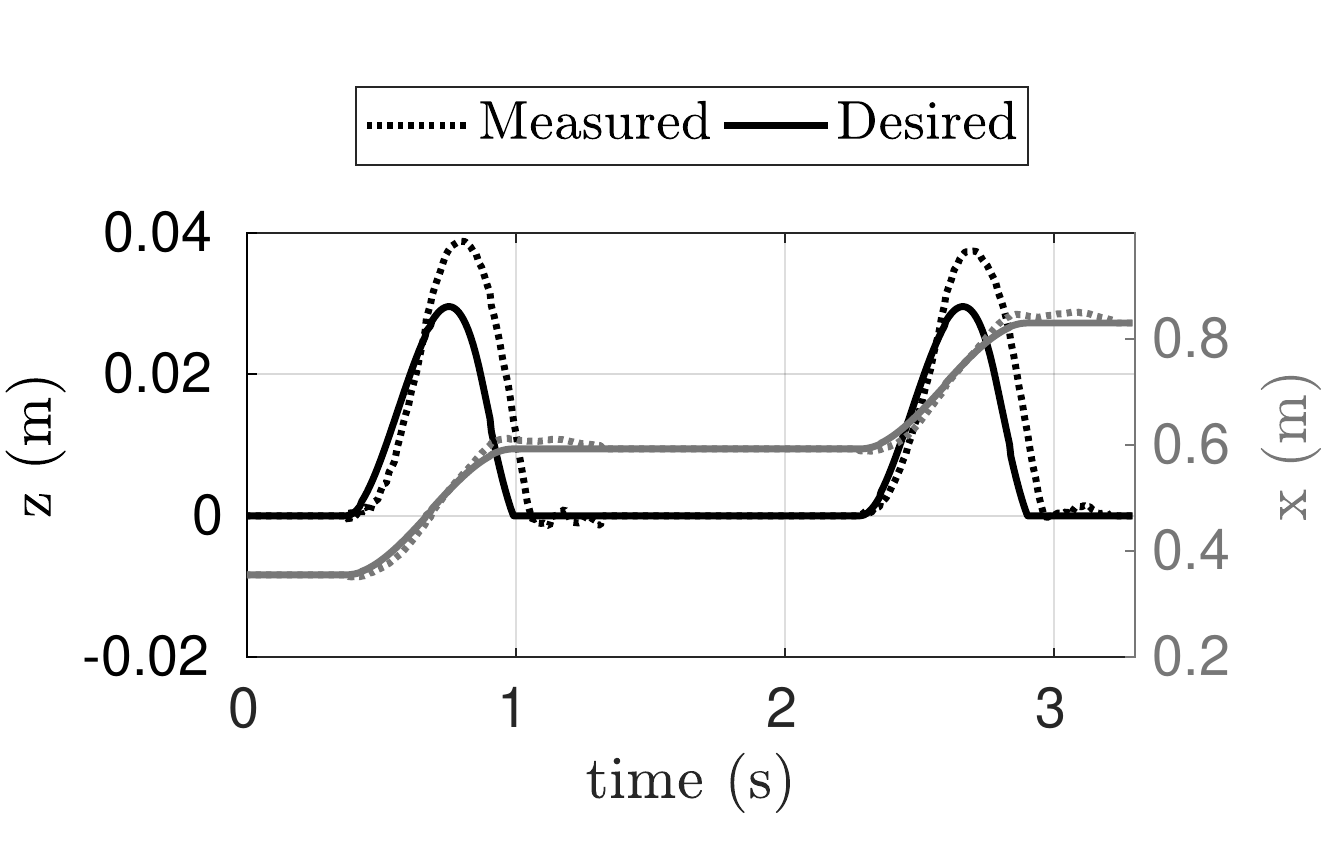}
        \caption{$\SI{0.1563}{\meter \per \second}$}
        \label{fig:inst_vel-min_vel-lf}
    \end{subfigure}
    \begin{subfigure}[t]{0.49\columnwidth}
        \centering
        \includegraphics[width=\textwidth]{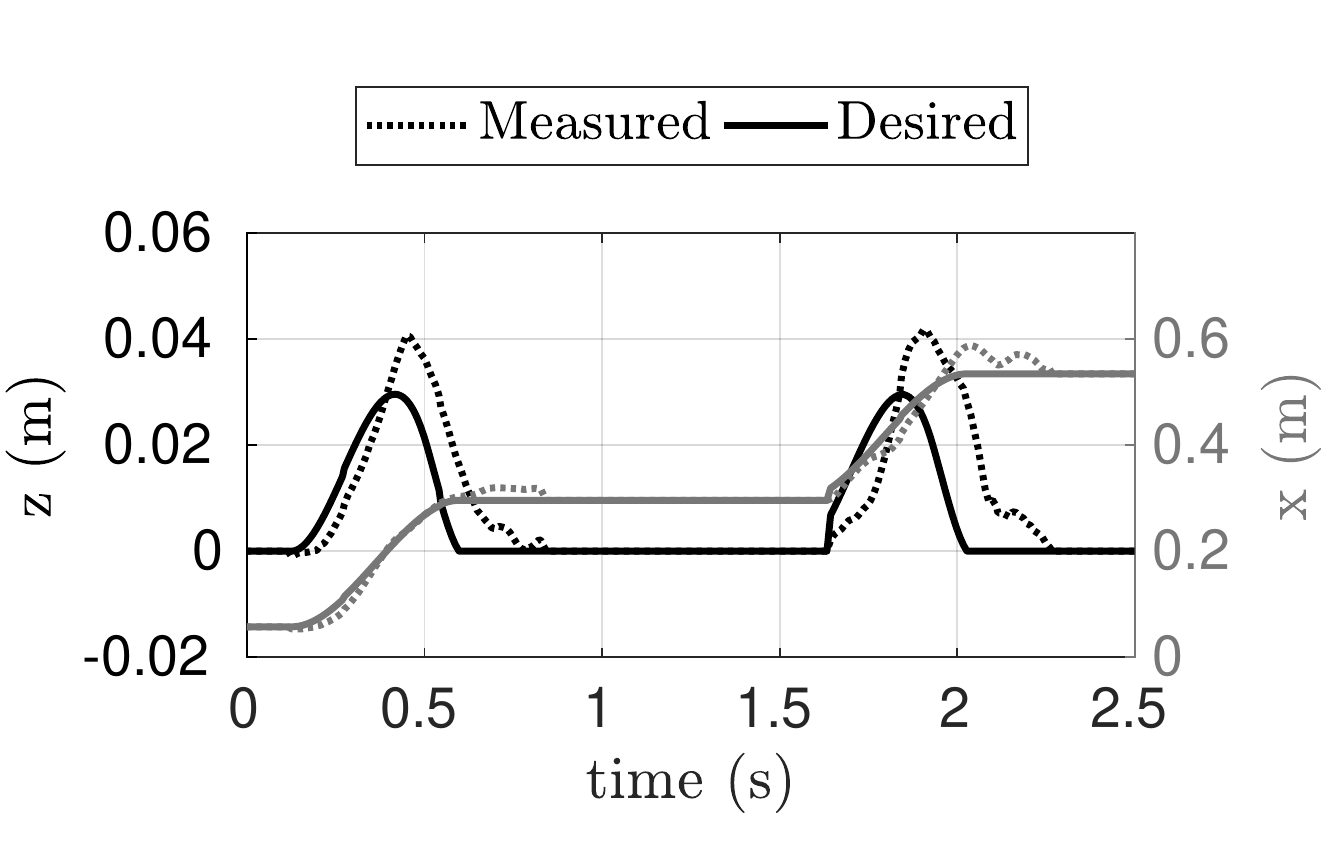}
        \caption{$\SI{0.3372}{\meter \per \second}$}
        \label{fig:inst_vel-max_vel-lf}
    \end{subfigure}
    \end{myframe}
        \end{subfigure}
     \caption{Tracking of the left foot position using the instantaneous simplified model control. Whole-body QP  control as position control (a) and  velocity control (c). Whole-body QP control as position control (b) and  velocity control (d), walking velocity.}
\end{figure}

\paragraph{Simplified model control: Predictive versus Instantaneous}
\label{sub_sec:Simplified Models Controller}
In this section, the control laws presented in Secs.~\ref{sec:instantaneous-control} and \ref{sec:predictive-control} are compared. To simplify the analysis, the implementation of the \emph{whole-body QP layer} is kept fixed, and only the results when the robot is in position control are discussed.

In Figs.~\ref{fig:inst_pos-min_vel-dcm} and \ref{fig:mpc_pos-min_vel-dcm}, the DCM tracking performances obtained with the instantaneous and predictive controllers are depicted. When the robot walks with a forward velocity of $\SI{0.1563}{\meter \per \second}$, both the implementations guarantee good tracking performances, with a DCM error below $\SI{5}{\centi \meter}$.
It is interesting to notice that the use of the instantaneous controller induces faster variations of the measured DCM and consequentially overall vibrations of the robot. One of the reasons for this variation is attributable to the desired ZMP injected by the controller. Indeed the desired ZMP is proportional to the measured DCM, that it is, in general, affected by noise. 
Filtering the DCM may be a possible solution. Nevertheless, our experience showed that the addition of a filter may degrade the controller performances due to the consequent introduction of delays.

Figs.~\ref{fig:inst_pos-min_vel-com} and \ref{fig:mpc_pos-min_vel-com} depict the CoM tracking performances when the walking velocity is $\SI{0.1563}{\meter \per \second}$. These performances are mainly dependent on the ZMP-CoM controller~\eqref{eq:zmp_controller}. This controller receives the desired CoM trajectory and desired ZMP values from either the instantaneous or predictive controllers. In both cases, the controller guarantees good tracking, with a maximum CoM error of $\SI{2}{\centi \meter}$. 

Figs.~\ref{fig:inst_pos-min_vel-zmp} and~\ref{fig:mpc_pos-min_vel-zmp} represent the ZMP tracking performances, which are still mainly dependent on the ZMP-CoM controller~\eqref{eq:zmp_controller}. It is important to observe that the desired ZMP is smoother when it is generated by using the predictive law. Indeed, this property is related to the associated weight in the cost function~\eqref{eq:mpc_cost} of the MPC problem. Although this smoother behavior does contribute to decrease the overall vibrations, the  desired ZMP is also bounded and therefore the system has less manoeuvrability than when the instantaneous controller is used. Thus the robot becomes less reactive. Despite the extensive hand-made tuning, we were not able to increase the velocity when the \emph{simplified model control} used the predictive law.
\par
Figs.~\ref{fig:mpc_pos-max_vel-dcm} and~\ref{fig:inst_pos-max_vel-dcm} depict the DCM tracking performances with the robot desired walking speed of $\SI{0.3372}{\meter \per \second}$. Initially ($t < \SI{1.5}{\second}$), there is no significant difference between the DCM tracking obtained with instantaneous and predictive control laws.
\par
However, around $t = \SI{1.5}{\second}$, the fast variation of the desired DCM induces the drop of the tracking performances. Consequentially this performances drop induces an overall bad tracking of the CoM and ZMP and at $t\approx \SI{2}{\second}$ the robot fall. In order to increase the responsiveness of the controller, one may increase the gains of the ZMP-CoM controller, however we notice that by increasing these gains, the system is more sensitive to the external disturbances and noise.
In a nutshell, the \emph{predictive simplified control} is much less robust than the \emph{instantaneous simplified control} with respect to ZMP tracking errors. To face this problem, we suggest increasing the gain $K_{zmp}$ of the ZMP-COM controller~\eqref{eq:zmp_controller}. However, to avoid injecting noise due to the force sensors we suggest to add a filter. In our case, we decide to avoid to use the low pass filter because the delay introduced dropped the overall performances.

\paragraph{Whole-Body QP Control: Position versus Velocity}
To simplify the analysis, the implementation of the \emph{simplified model control} is kept fixed, and only the results with the instantaneous control~\eqref{eq:instanteneous_dcm} are presented.
Furthermore, for the sake of compactness, we decide to present only the tracking of the desired feet positions, similar considerations hold for the tracking of the CoM.
\par
Figs. \ref{fig:inst_pos-min_vel-lf} and \ref{fig:inst_vel-min_vel-lf} depict the tracking of desired left foot positions when the robot is in position and velocity controlled, respectively. The position controller ensures better tracking performance than the velocity one. One may consider increasing the gains of the controllers~\eqref{eq:velocity_control_jacobians}, however increasing too much the gains induces overall oscillation in the robot.
\par
The aforementioned foot tracking problem worsens at higher walking velocity. Fig.~\ref{fig:inst_pos-max_vel-lf} shows that the feet tracking error is lower than $\SI{5}{\centi \meter}$ on the $x$ axis and $\SI{0.5}{\centi \meter}$ on the $z$ one for position control. Instead, the velocity control in Fig.~\ref{fig:inst_vel-max_vel-lf} keeps the error always lower than $\SI{6}{\centi \meter}$ and $\SI{3}{\centi \meter}$ on the the $x$ and $y$ components, respectively.

\subsubsection{Dynamics-based Walking Architecture}
Controlling the robot using a torque controller architecture is not an easy task.
Indeed, the performance guaranteed by the position/velocity architecture are not reached because of an imperfect low-level torque control, presence of friction and model errors.
For this reason, to validate the torque architecture, we decide to present also the simulation results.
When the robot is torque controlled, the noise affecting the measured DCM does not allow us to use the simplified model controllers. Thus we decided to stabilize a desired CoM instead of DCM.
Indeed the simplified model control, either the instantaneous~\ref{sec:instantaneous-control} or the predictive controller~\ref{sec:predictive-control}, injects a (desired) ZMP that depends on the measured DCM. As the consequence, it generates undesired vibrations on the robot. We also tried to implement low pass filters for mitigating such behavior. However, we did not find the right trade-off for obtaining overall performance improvements.
Although the extensive hand-made tuning of the simplified model controllers, we were not able to close the loop on the desired DCM. Tracking down the source of the DCM noise to the measured joint velocities, we decided to stabilize a desired CoM trajectory instead. In order to maintain consistency with the previous architectures, we generate such trajectory from the LIPM dynamics~\eqref{eq:simplified-model} starting from a desired DCM trajectory.
\par
\begin{figure}[t]
    \begin{myframe}{Torque Control}
        \begin{subfigure}[b]{0.49\textwidth}
        \centering
        \includegraphics[width=\textwidth]{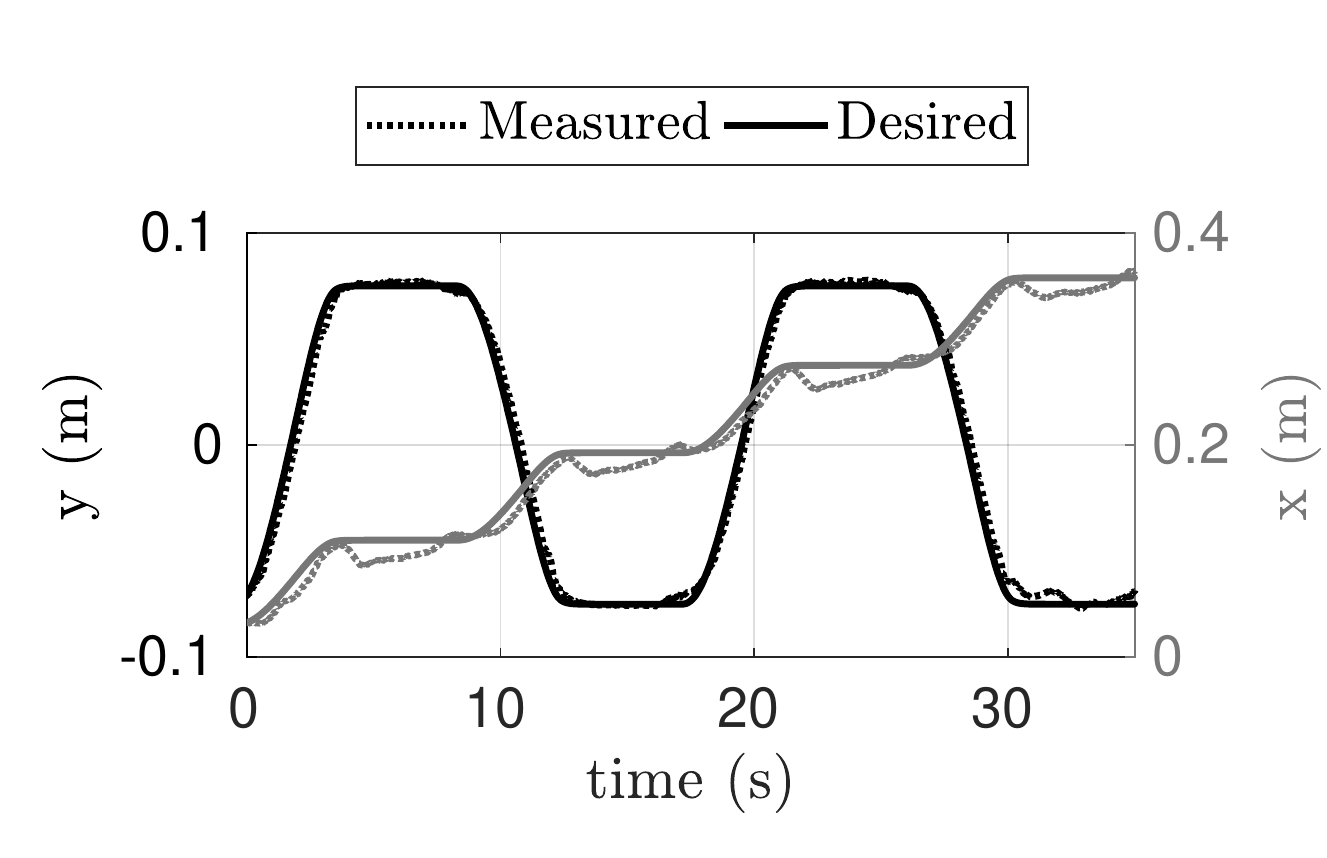}
        \caption{CoM}
        \label{fig:torq-real-com}
    \end{subfigure}
    \hfill
     \begin{subfigure}[b]{0.49\textwidth}
        \centering
        \includegraphics[width=\textwidth]{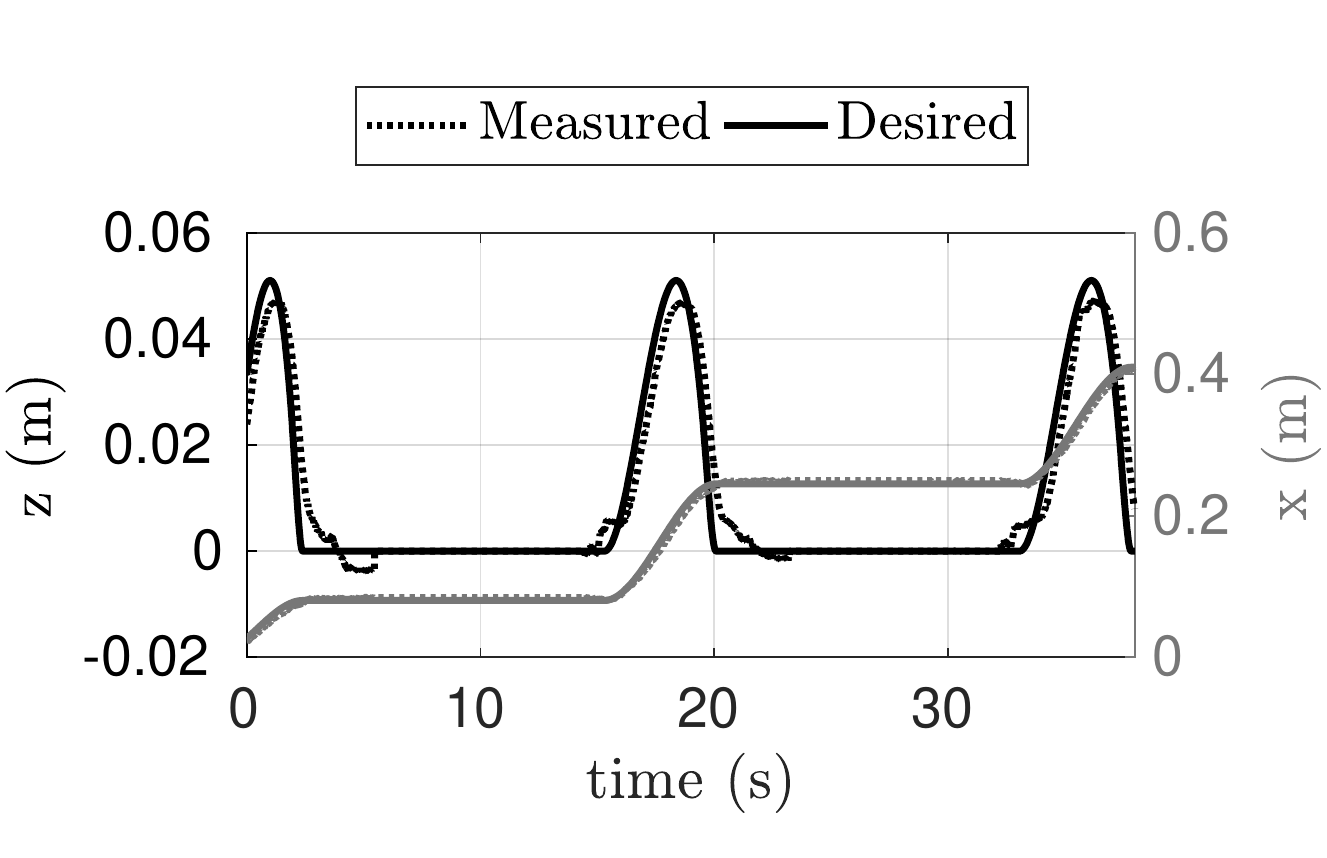}
        \caption{Left foot}
        \label{fig:torq-real-lf}
    \end{subfigure}
    \end{myframe}
    \caption{Tracking of the CoM (a), and left foot position (b) with whole-body QP control as torque control.}
\end{figure}

Table~\ref{tab:max_velocity_trq} summarizes the maximum velocities achieved using different implementations of the dynamics-based architecture. The labels \emph{simulation} and \emph{real robot} mean that the experiments are carried out on the Gazebo Simulator \cite{koenig2004design} or the real platform, respectively.
\begin{table}[t]
\centering
\tbl{Maximum forward walking velocities achieved in simulation and \\ in a real scenario in case of torque controlled robot. \label{tab:max_velocity_trq}}
{\begin{tabular}{@{}cc|c@{}} \toprule
Platform & Simplified Model Control & Max Straight Velocity (m/s) \\
\colrule
Real Robot & -  &  0.0186\\
Simulation & Instantaneous  & 0.2120\\
Simulation & Predictive  &  0.1448\\
\botrule
\end{tabular}}
\end{table}
\paragraph{Experiments on the Real Robot}
In this section, we present the performance of the walking architecture when the robot is torque controlled. 
Fig.~\ref{fig:torq-real-com} depicts the CoM tracking performances. It is important to notice that the tracking error on the x-axis is greater than the one on the y-axis. To reduce this, one may tend to increase the associated gain. However, our experience showed that increasing the CoM gain contributes to the overall vibration of the robot.
\par
Fig.~\ref{fig:torq-real-lf} depicts the tracking of the desired left foot trajectory. Event if the walking velocity is lower than the one used for the kinematics based architecture, the dynamics based whole-body QP is not able to guarantee good performances. One may consider increasing the gains of the feet controller~\eqref{eq:torque_feet_linear_pid}, although the extensive hand-made tuning, we were not able to increase the robot velocity.

\par
\begin{figure}[t]
     \begin{myframe}{Torque Control}
    \begin{subfigure}[b]{0.327\textwidth}
        \centering
        \includegraphics[width=\textwidth]{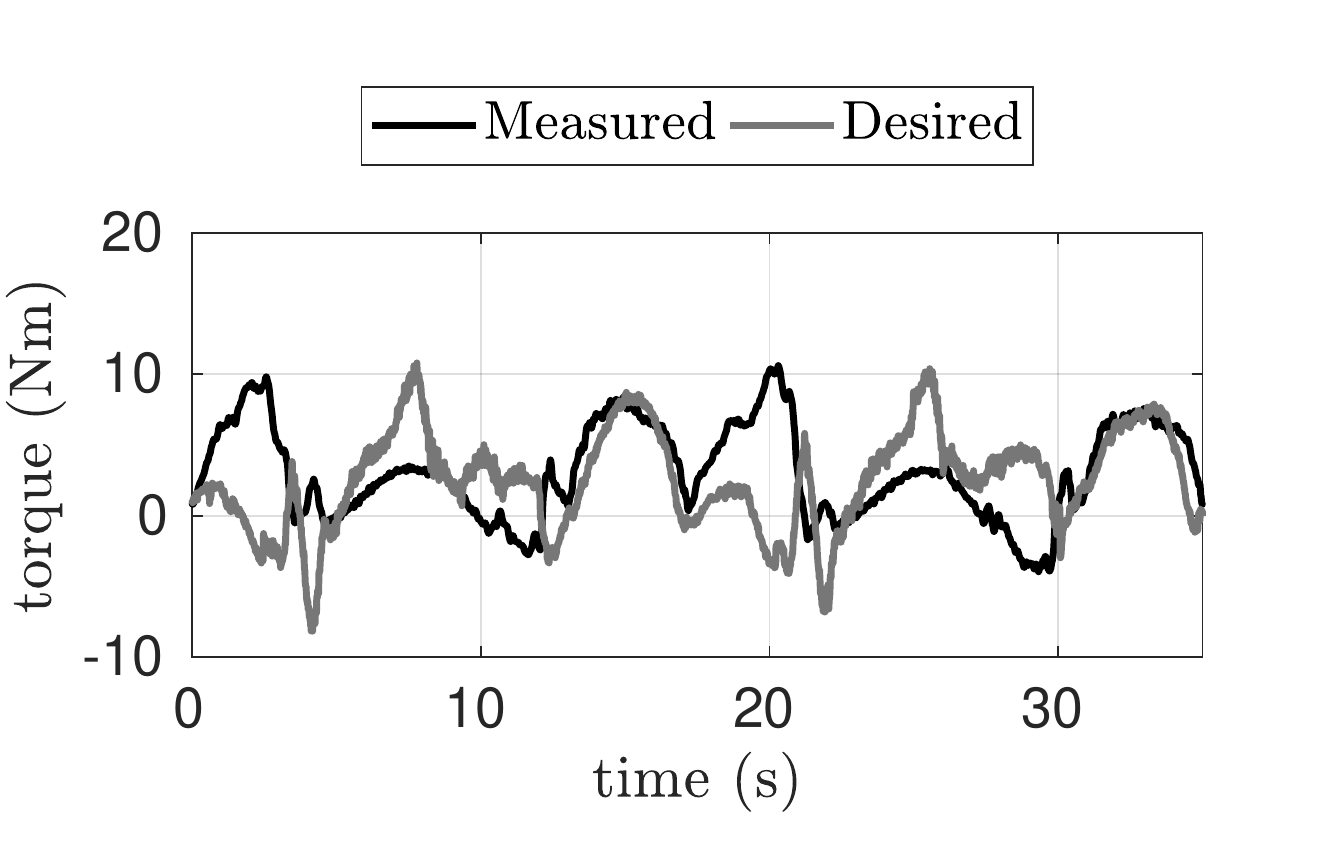}
        \caption{Hip pitch}
        \label{fig:real-hip_pitch_trq}
    \end{subfigure}
    \hfill
    \begin{subfigure}[b]{0.327\textwidth}
        \centering
        \includegraphics[width=\textwidth]{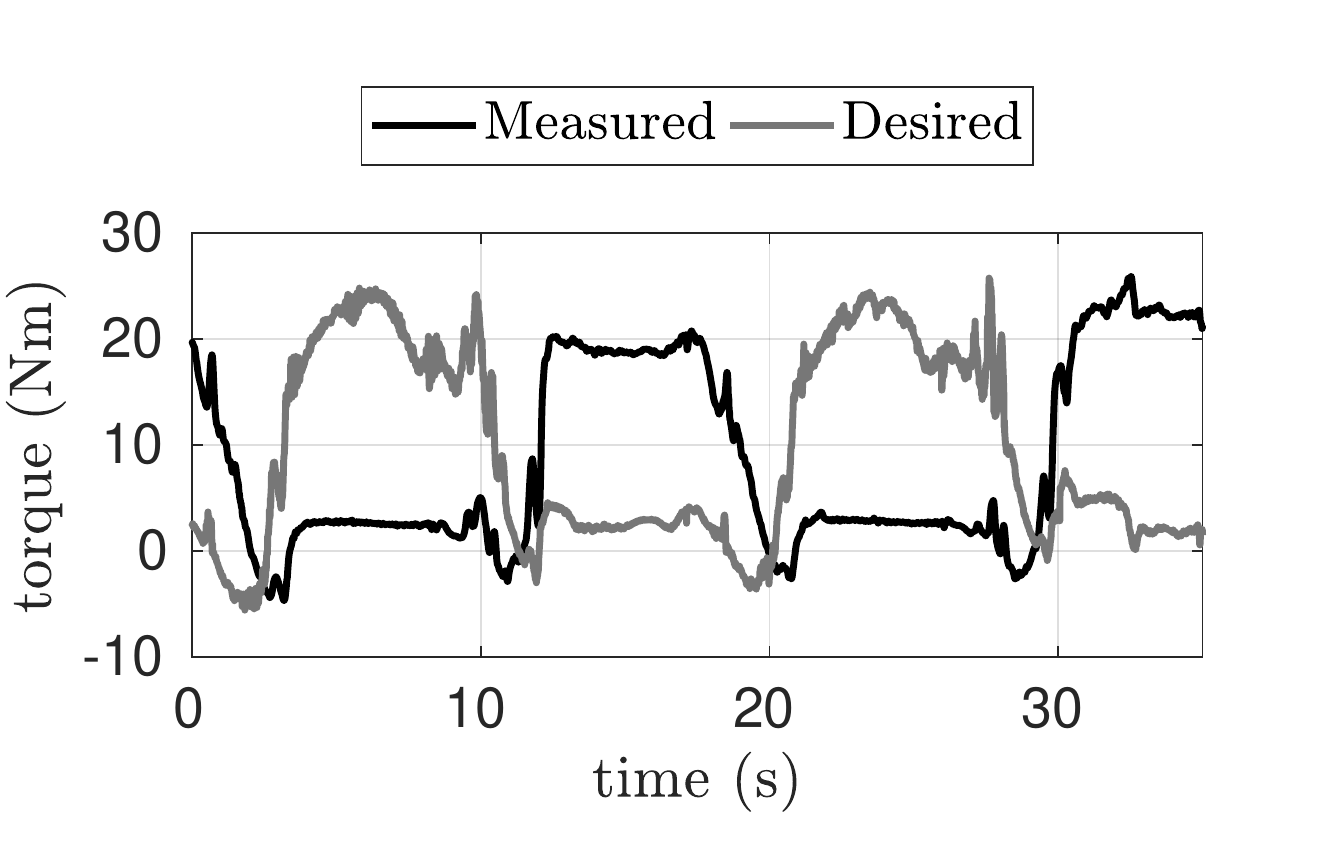}
        \caption{Hip roll}
        \label{fig:real-hip_roll_trq}
    \end{subfigure} 
    \hfill
    \begin{subfigure}[b]{0.327\textwidth}
        \centering
        \includegraphics[width=\textwidth]{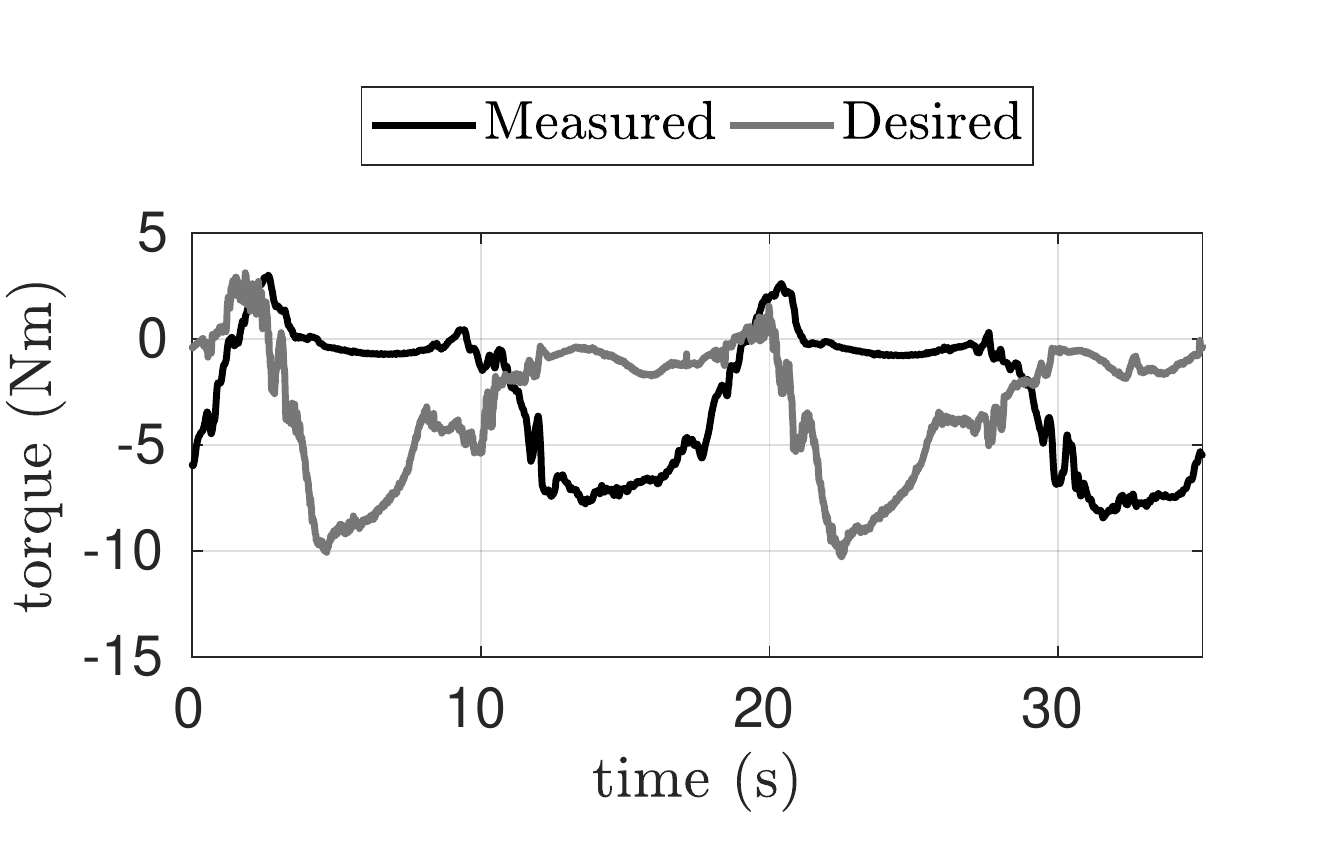}
        \caption{Hip yaw}
        \label{fig:real-hip_yaw_trq}
    \end{subfigure}
    \begin{subfigure}[b]{0.327\textwidth}
        \centering
        \includegraphics[width=\textwidth]{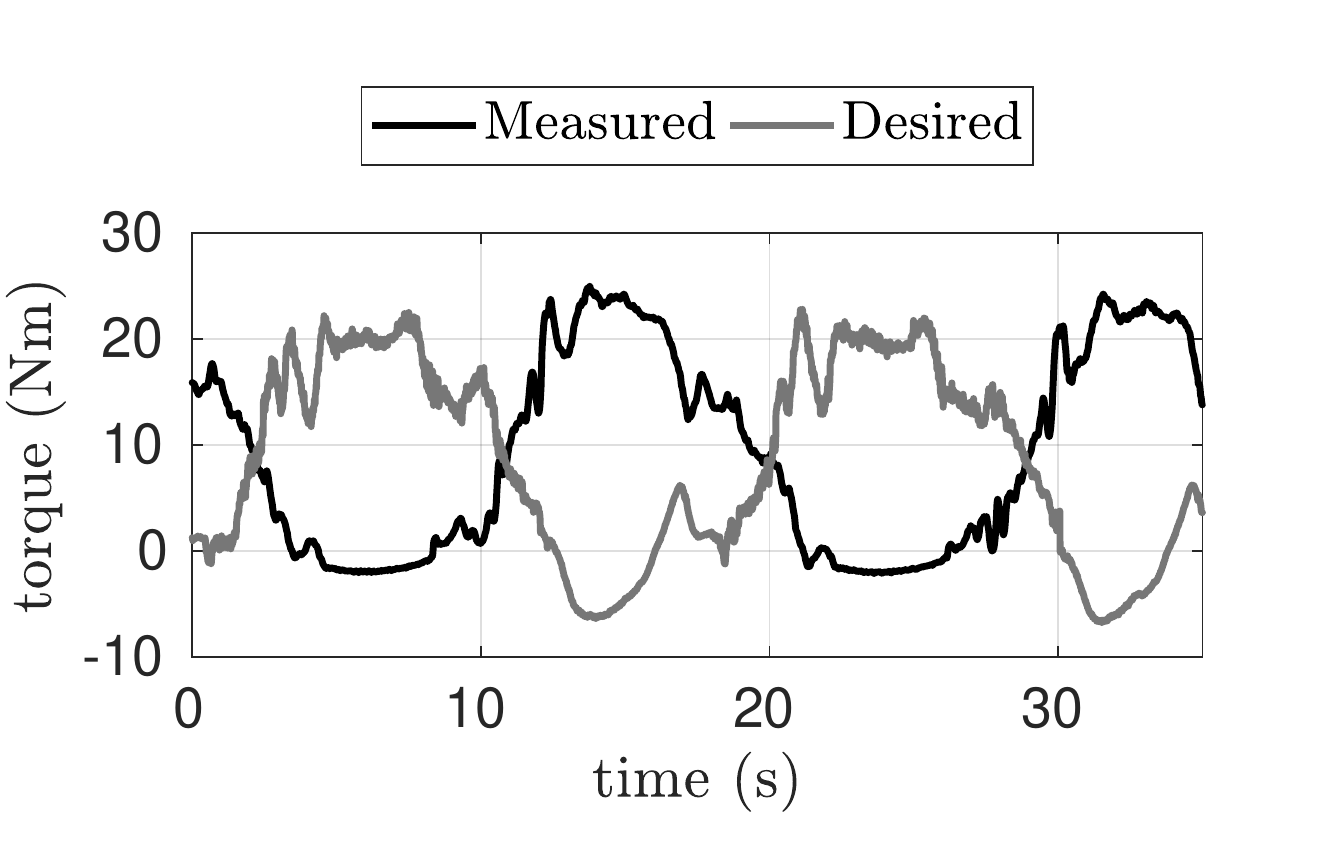}
        \caption{Knee}
        \label{fig:real-knee_trq}
    \end{subfigure} 
    \hfill
        \begin{subfigure}[b]{0.327\textwidth}
        \centering
        \includegraphics[width=\textwidth]{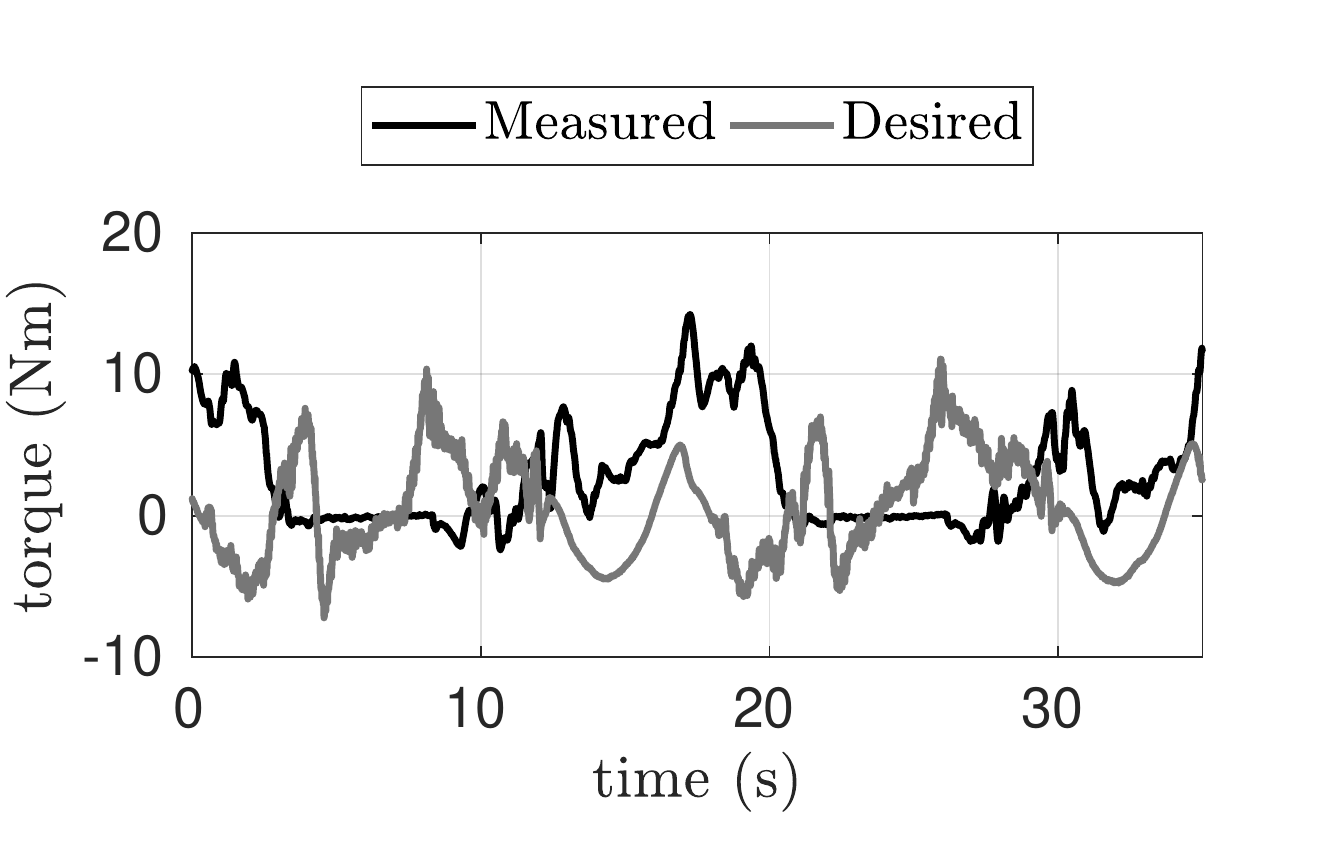}
        \caption{Ankle pitch}
        \label{fig:real-ankle_pitch_trq}
    \end{subfigure} 
    \begin{subfigure}[b]{0.327\textwidth}
        \centering
        \includegraphics[width=\textwidth]{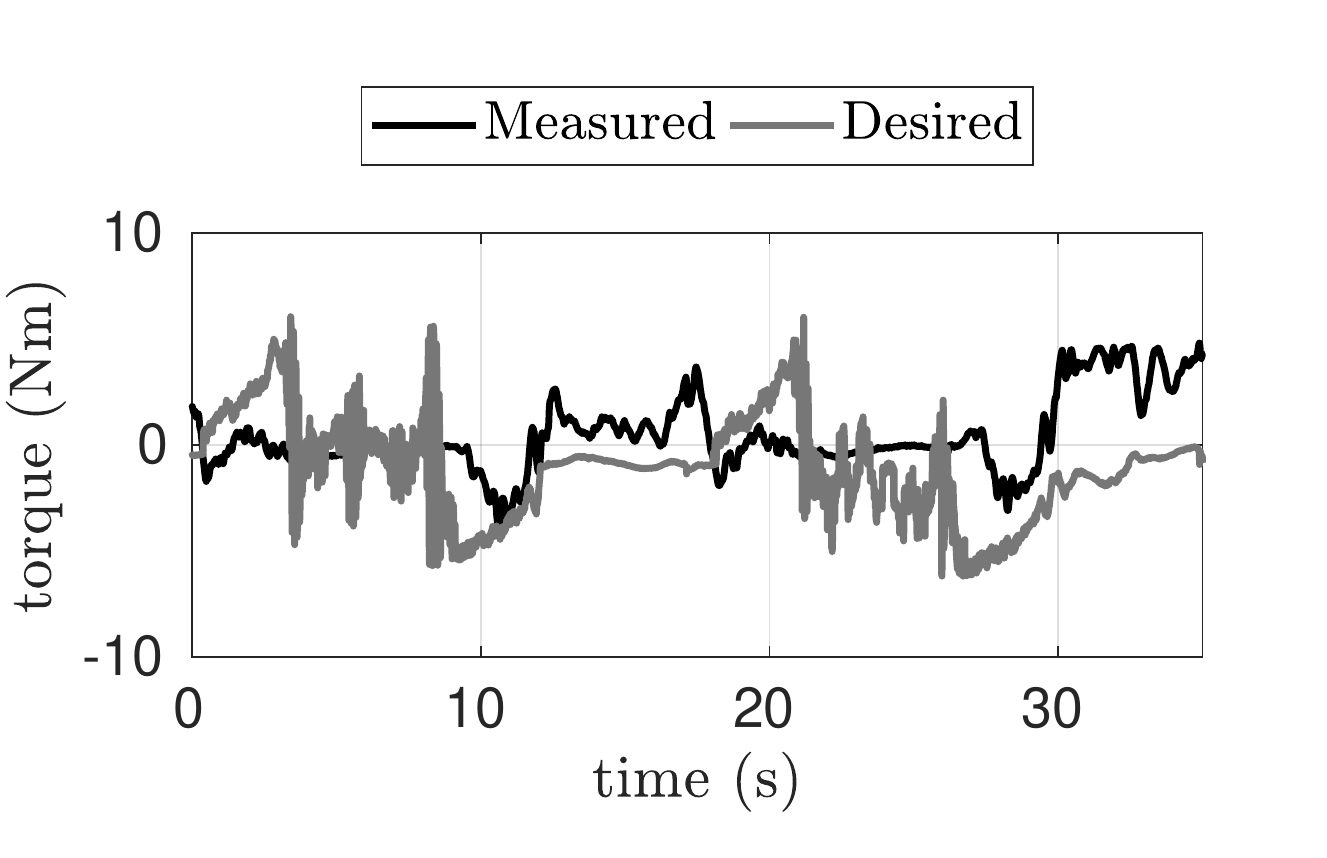}
        \caption{Ankle roll}
        \label{fig:real-ankle_roll_trq}
    \end{subfigure}
    \end{myframe}
    \caption{Tracking of the desired joint torques of the left leg.}
    \label{fig:joint_trq}
\end{figure}

Such bad performances may be attributed to the low-level torque controller. Indeed as depicted in Fig.~\ref{fig:joint_trq} the tracking performances of the low-level torque control are poor. One is tempted to increase the gains of the low-level torque controller for ensuring better performances. However, since the iCub robot does not have joint torque sensors, the joint torques are estimated by using the readouts of the force-torque sensors. We observed that the noise due to the force-torque sensors is harmful to the estimated torque and, consequentially increasing too much the gains causes undesired overall vibrations.
\paragraph{Experiments on the Simulation Scenario}
In this section, we present the simulation results. To simplify the analysis we decide to show only the results when the robot walks with a forward velocity of $\SI{0.1448}{\meter \per \second}$. 
\par
\begin{figure}[t]
         \begin{subfigure}[b]{0.66\textwidth}    
     \begin{myframe}{Instantaneous}
    \begin{subfigure}[b]{0.49\textwidth}
        \centering
        \includegraphics[width=\textwidth]{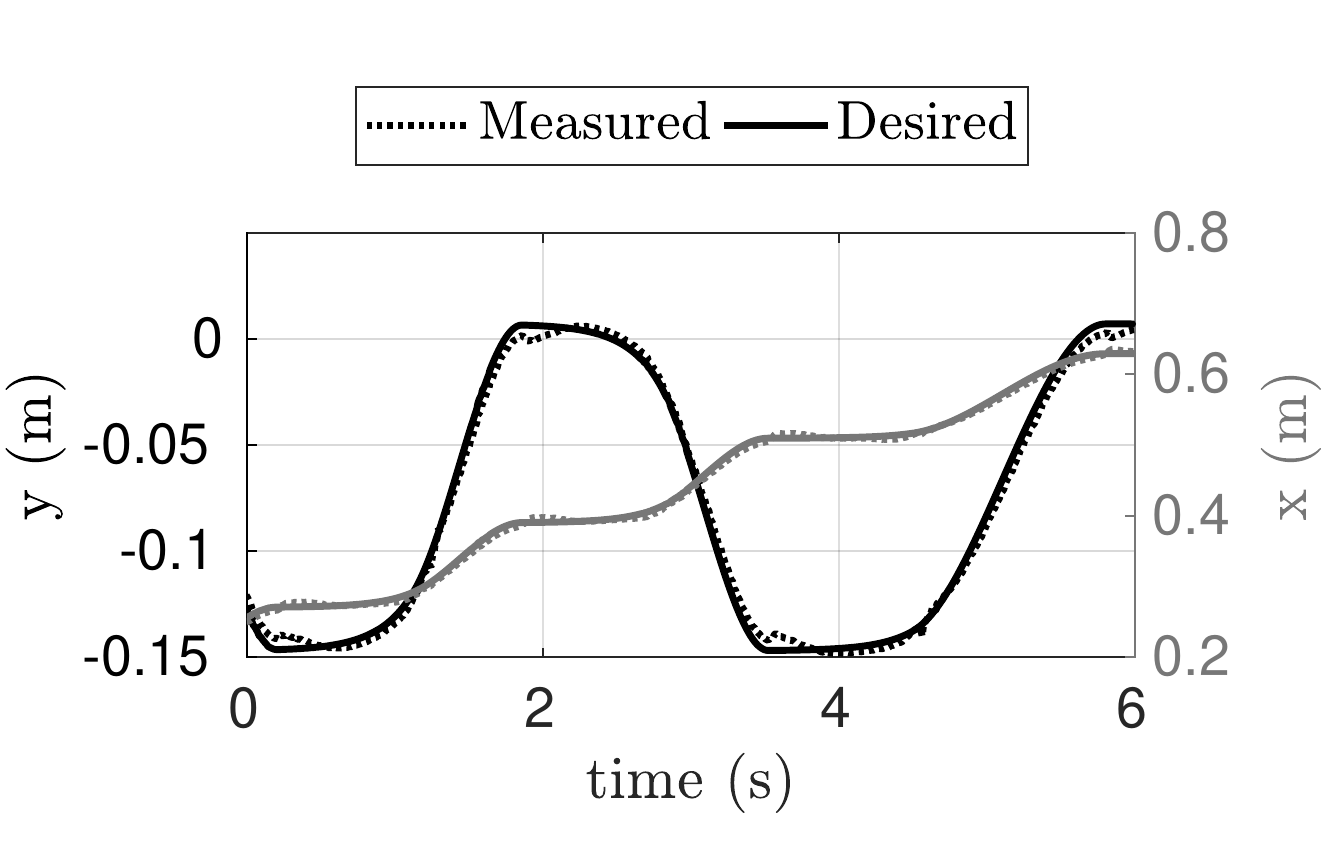}
        \caption{DCM}
        \label{fig:inst_torq_sim-min_vel-dcm}
    \end{subfigure}
    \begin{subfigure}[b]{0.49\textwidth}
        \centering
        \includegraphics[width=\textwidth]{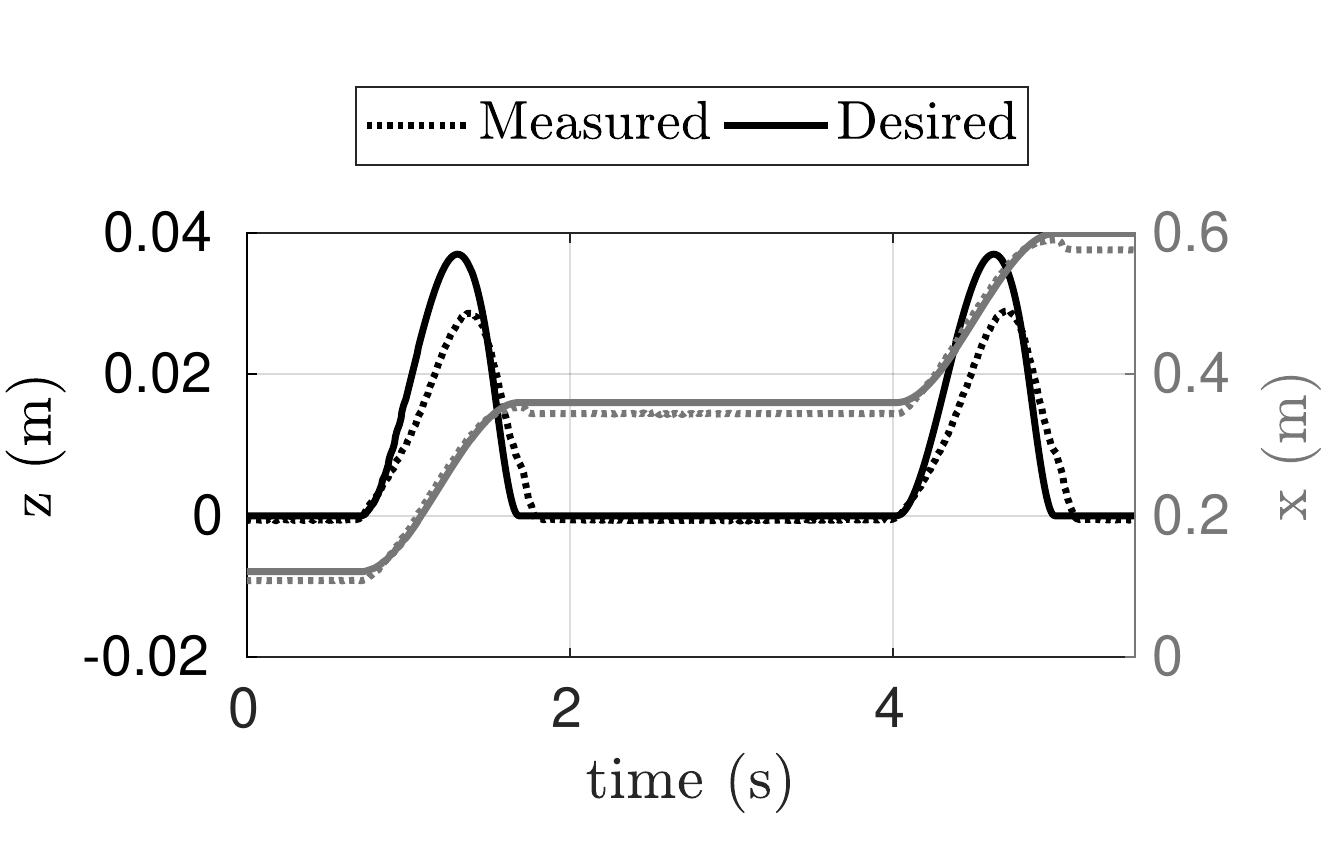}
        \caption{Left foot}
        \label{fig:inst_torq_sim-min_vel-lf}
    \end{subfigure} 
    \end{myframe}
    \end{subfigure} 
    \begin{subfigure}[b]{0.333\textwidth}    
    \begin{myframe}{Predictive}
    \begin{subfigure}[b]{1\textwidth}
        \centering
        \includegraphics[width=\textwidth]{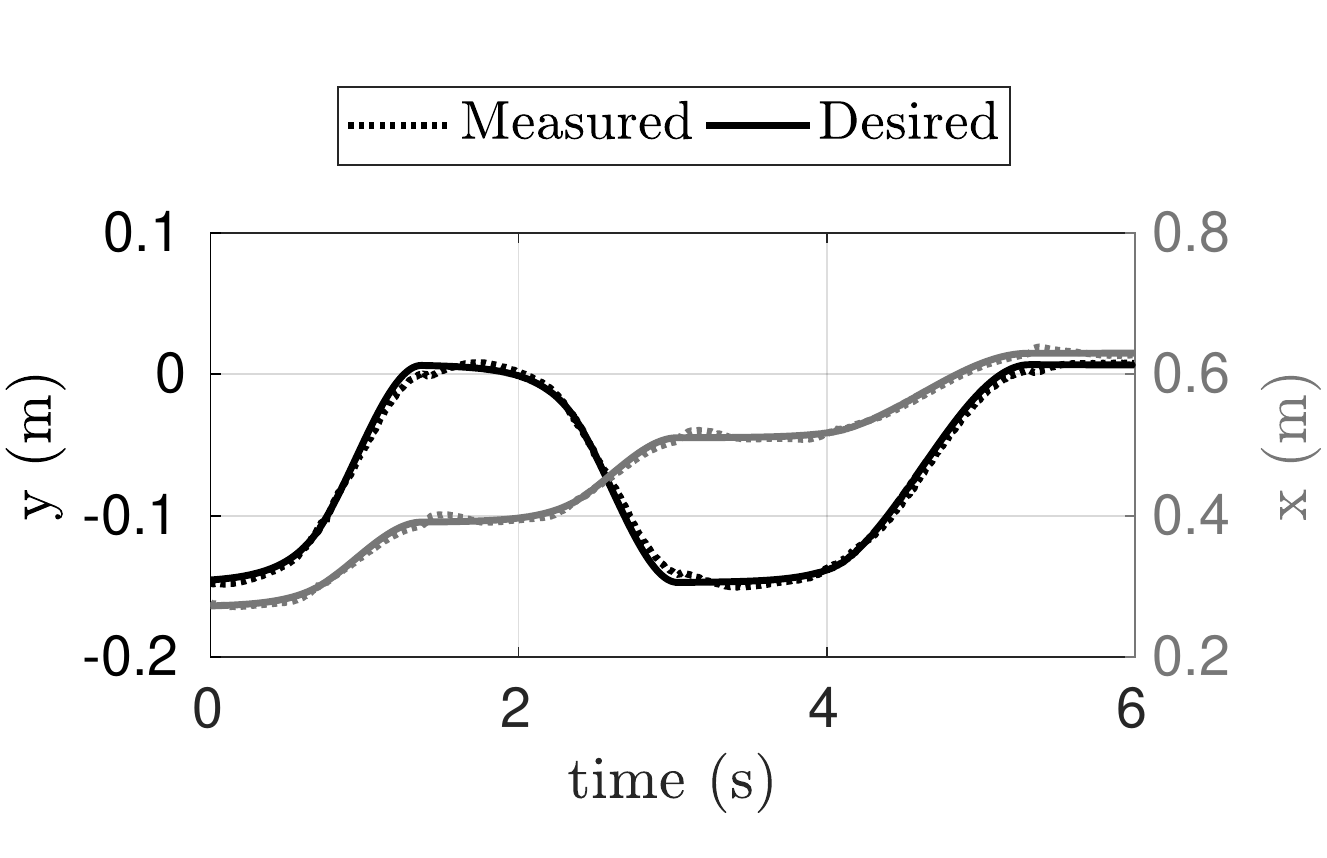}
        \caption{DCM}
        \label{fig:mpc_torq_sim-min_vel-dcm}
    \end{subfigure}
    \end{myframe}
    \end{subfigure}
    \caption{Tracking of the left foot position using the instantaneous simplified model control and Whole-body QP control as torque control (b). Tracking of the DCM trajectory using the instantaneous (a) and predictive (c) simplified model controls. Forward velocity: $\SI{0.1448}{\meter\per\second}$.}
\end{figure}

Figs~\ref{fig:inst_torq_sim-min_vel-dcm} and \ref{fig:mpc_torq_sim-min_vel-dcm} depicts the tracking performance with the instantaneous and the predictive controller, respectively. Both implementations guarantee excellent performances, with a DCM error below $\SI{1}{\centi \meter}$. 
Notice that when the simplified model controller layer is implemented with the instantaneous controller, the whole-body QP control layer sometimes fails to find an admissible solution. This happens because the desired ZMP, evaluated using the instantaneous controller, may exit the feet support polygon, so it may be not feasible. To face this issue we suggest projecting the desired ZMP onto the support polygon.\cite{Englsberger2011}
\par
Fig.~\ref{fig:inst_torq_sim-min_vel-lf} depicts the tracking of the desired left foot trajectory.  The controller is able to guarantee a tracking error always below $\SI{1}{\centi\meter}$.

\subsection{Energy consumption}
To compare the energy efficiency of different control architecture we use the \emph{Specific Energetic Cost}. The \emph{Specific Energetic Cost} is defined as: \cite{Torricelli}
\begin{equation}
    c_{et} = \frac{E}{m D},
\end{equation}
where $E$ is the positive mechanical work of the actuation system, $m$ is the mass of the system and $D$ is the distance traveled. 
\par
Table~\ref{tab:energy_consumption} summarizes the Specific Energetic Cost evaluated using different implementations of the architecture. The labels \emph{Simulation} and \emph{Real Robot} mean that the experiments are carried out on the Gazebo Simulator or the real platform, respectively. The labels, \emph{Position} and \emph{Torque} control, instead, mean that the whole-body QP control layer outputs are either desired joint positions or torque, respectively. 
We noticed the Dynamics-based architecture has a lower Specific Energetic Cost \todo[inline]{ST: I am afraid that "is more efficient" is a big ambiguous and could lead to criticism. I would simply write a more neutral "has a lower Specific Energetic Cost".} than the Kinematics-based architecture, the reason of this result is attributable to the minimization of the joint torque when the robot is torque controlled -- see~\eqref{eq:torque_control_cost_torque}.
\par
\textbf{Remark} Let us observe that the \emph{Specific Energetic Cost} is one of the possible criteria for energy benchmarking. Other criteria, like electrical power, shall be considered to have a clear picture of energy expenditure during robot bipedal locomotion.

\begin{table}[t]
\centering
\tbl{Specific Energetic Cost evaluated in simulation and in a real scenario in case of torque and position controlled robot. \label{tab:energy_consumption}}
{\begin{tabular}{ccc|c} \toprule
Platform & Whole-Body QP Control & Velocity (m/s) & Specific Energetic Cost (J/Kg/m)\\
\colrule
Real Robot & Position & 0.0186  &  9.66\\
Real Robot & Torque & 0.0186  &  3.85\\
Simulation & Position & 0.2120  & 4.82\\
Simulation & Torque & 0.2120  &  2.55\\
\botrule
\end{tabular}}
\end{table}
\section{Conclusions and Future Work}
\label{sec:conclusions_and_future_work}
This paper contributes towards the benchmarking of different implementations of state-of-the-art control architectures for humanoid robots locomotion. In particular, we proposed a three-layers controller architecture, which exploits the concept of the Divergent Component of Motion. 
The cornerstone of this work is the comparison of the different implementations of the three layers, namely the trajectory optimization, the simplified model control and the whole-body QP control.
In particular, for the simplified model control layer, we discussed the results obtained with the predictive and instantaneous controllers implementation. Furthermore, we compare the performances obtained controlling the robot in position, velocity and torques modes.
Even if the proposed controller architecture was tested only on the humanoid robot iCub,\cite{Natale2017} it can be also used for other typology of structures.\cite{russo,Collins}
\par
We show that instantaneous controllers combined with robot position control allowed us to achieve a desired walking speed of $\SI{0.3372}{\meter \per \second}$.
On the other hand, we show that the torque-based architecture produces excellent results in simulation. Unstructured uncertainty on the real robot caused an overall performance degradation in the real scenario. 
Furthermore analyzing the energy consumption, we show that when the robot is torque-controlled the specific energy cost required to walk is lower than in a position-controlled robot.
\par
In summary, we demonstrate that instantaneous controllers coupled with robot position control enabled us to reach the highest walking velocity ever achieved on the humanoid robot iCub. The torque-based architecture, on the other hand, generates impressive simulation outcomes, however, uncertainties on the real robot make difficult to use a torque-control architecture in a real scenario.
In both simulation and real scenario, when the robot is torque-controlled, the specific energy cost required to walk is lower than in a position-controlled robot. As a consequence, even if the Dynamics-based architecture does not allow iCub to walk at the same velocity reached with the Kinematics-based architecture, it is always more efficient because requires less control effort.
\par
As future work, we intend to improve the walking performances when the robot is in torque mode. To achieve this goal we plan to improve the tracking performances of the low-level torque controller by using the current for estimating the joint torques.\cite{Zhang2015} We also plan to develop a whole-body torque control architecture that takes advantage of the joints elasticity of the humanoid robot iCub (e.g.\cite{Hopkins_ijhr}).
Another interesting future work is the implementation of a footstep adjustment algorithm.\cite{Griffin2016,Shafiee-Ashtiani2017,shafiee2019online} This will increase the overall robustness in case of large disturbances acting on the robot.

\bibliography{Locomotion-papers.bib}

\begin{thebibliography}{10}
\providecommand{\url}[1]{#1}
\csname url@rmstyle\endcsname
\providecommand{\newblock}{\relax}
\providecommand{\bibinfo}[2]{#2}
\providecommand\BIBentrySTDinterwordspacing{\spaceskip=0pt\relax}
\providecommand\BIBentryALTinterwordstretchfactor{4}
\providecommand\BIBentryALTinterwordspacing{\spaceskip=\fontdimen2\font plus
\BIBentryALTinterwordstretchfactor\fontdimen3\font minus
  \fontdimen4\font\relax}
\providecommand\BIBforeignlanguage[2]{{%
\expandafter\ifx\csname l@#1\endcsname\relax
\typeout{** WARNING: IEEEtran.bst: No hyphenation pattern has been}%
\typeout{** loaded for the language `#1'. Using the pattern for}%
\typeout{** the default language instead.}%
\else
\language=\csname l@#1\endcsname
\fi
#2}}

\bibitem{feng2015optimization}
S.~Feng, E.~Whitman, X.~Xinjilefu, and C.~G. Atkeson, ``{Optimization-based
  Full Body Control for the DARPA Robotics Challenge},'' \emph{J. F. Robot.},
  vol.~32, no.~2, pp. 293--312, 2015.

\bibitem{dai2014whole}
H.~Dai, A.~Valenzuela, and R.~Tedrake, ``{Whole-body motion planning with
  centroidal dynamics and full kinematics},'' in \emph{2014 IEEE-RAS Int. Conf.
  Humanoid Robot.}\hskip 1em plus 0.5em minus 0.4em\relax IEEE, 2014, pp.
  295--302.

\bibitem{herzog2015trajectory}
A.~Herzog, N.~Rotella, S.~Schaal, and L.~Righetti, ``{Trajectory generation for
  multi-contact momentum control},'' in \emph{Humanoid Robot. (Humanoids), 2015
  IEEE-RAS 15th Int. Conf.}\hskip 1em plus 0.5em minus 0.4em\relax IEEE, 2015,
  pp. 874--880.

\bibitem{PascalHandbook}
P.~Morin and C.~Samson, \emph{{Handbook of Robotics}}.\hskip 1em plus 0.5em
  minus 0.4em\relax Springer, 2008, ch. Motion con, pp. 799--826.

\bibitem{flavigne2010reactive}
D.~Flavigne, J.~Pettr{\'{e}}e, K.~Mombaur, J.-P. Laumond, and Others,
  ``{Reactive synthesizing of human locomotion combining nonholonomic and
  holonomic behaviors},'' in \emph{Biomed. Robot. Biomechatronics (BioRob),
  2010 3rd IEEE RAS EMBS Int. Conf.}\hskip 1em plus 0.5em minus 0.4em\relax
  IEEE, 2010, pp. 632--637.

\bibitem{8594277}
S.~Dafarra, G.~Nava, M.~Charbonneau, N.~Guedelha, F.~Andradel, S.~Traversaro,
  L.~Fiorio, F.~Romano, F.~Nori, G.~Metta, and D.~Pucci, ``{A Control
  Architecture with Online Predictive Planning for Position and Torque
  Controlled Walking of Humanoid Robots},'' in \emph{2018 IEEE/RSJ Int. Conf.
  Intell. Robot. Syst.}, 2018, pp. 1--9.

\bibitem{Kajita2001}
S.~Kajita, F.~Kanehiro, K.~Kaneko, K.~Yokoi, and H.~Hirukawa, ``{The 3D linear
  inverted pendulum model: a simple modeling for biped walking pattern
  generation},'' \emph{Proc. 2001 IEEE/RSJ Int. Conf. Intell. Robot. Syst.},
  no. October 2016, pp. 239--246, 2001.

\bibitem{Pratt2006}
J.~Pratt, J.~Carff, S.~Drakunov, and A.~Goswami, ``{Capture point: A step
  toward humanoid push recovery},'' in \emph{Proc. 2006 6th IEEE-RAS Int. Conf.
  Humanoid Robot. Humanoids}, 2006, pp. 200--207.

\bibitem{Vukobratovic1969}
M.~Vukobratovic and D.~Juricic, ``{Contribution to the Synthesis of Biped
  Gait},'' \emph{IEEE Trans. Biomed. Eng.}, vol. BME-16, no.~1, pp. 1--6, 1969.

\bibitem{Kajita2003}
S.~Kajita, F.~Kanehiro, K.~Kaneko, K.~Fujiwara, K.~Harada, K.~Yokoi, and
  H.~Hirukawa, ``{Biped walking pattern generation by using preview control of
  zero-moment point},'' \emph{2003 IEEE Int. Conf. Robot. Autom. (Cat.
  No.03CH37422)}, vol.~2, no. October, pp. 1620--1626, 2003.

\bibitem{diedam2008online}
H.~Diedam, D.~Dimitrov, P.-B. Wieber, K.~Mombaur, and M.~Diehl, ``{Online
  walking gait generation with adaptive foot positioning through linear model
  predictive control},'' in \emph{2008 IEEE/RSJ Int. Conf. Intell. Robot.
  Syst.}\hskip 1em plus 0.5em minus 0.4em\relax IEEE, 2008, pp. 1121--1126.

\bibitem{Englsberger2015}
J.~Englsberger, C.~Ott, and A.~Albu-Sch{\"{a}}ffer, ``{Three-Dimensional
  Bipedal Walking Control Based on Divergent Component of Motion},'' \emph{IEEE
  Trans. Robot.}, vol.~31, no.~2, pp. 355--368, 2015.

\bibitem{Hopkins2015}
M.~A. Hopkins, D.~W. Hong, and A.~Leonessa, ``{Humanoid locomotion on uneven
  terrain using the time-varying divergent component of motion},'' in
  \emph{IEEE-RAS Int. Conf. Humanoid Robot.}, vol. 2015-Febru.\hskip 1em plus
  0.5em minus 0.4em\relax IEEE, nov 2015, pp. 266--272.

\bibitem{Stephens2010}
B.~J. Stephens and C.~G. Atkeson, ``{Dynamic Balance Force Control for
  compliant humanoid robots},'' in \emph{Intell. Robot. Syst. (IROS), 2010
  IEEE/RSJ Int. Conf.}, 2010, pp. 1248--1255.

\bibitem{nava16}
G.~Nava, F.~Romano, F.~Nori, and D.~Pucci, ``{Stability Analysis and Design of
  Momentum-based Controllers for Humanoid Robots},'' \emph{Intell. Robot. Syst.
  2016. IEEE Int. Conf.}, 2016.

\bibitem{Lee2016}
Y.~Lee, S.~Hwang, and J.~Park, ``{Balancing of humanoid robot using contact
  force/moment control by task-oriented whole body control framework},''
  \emph{Auton. Robots}, vol.~40, no.~3, pp. 457--472, mar 2016.

\bibitem{Feng2015a}
S.~Feng, X.~Xinjilefu, C.~G. Atkeson, and J.~Kim, ``{Optimization based
  controller design and implementation for the Atlas robot in the DARPA
  Robotics Challenge Finals},'' in \emph{IEEE-RAS Int. Conf. Humanoid Robot.},
  2015.

\bibitem{Kuindersma2016}
S.~Kuindersma, R.~Deits, M.~Fallon, A.~Valenzuela, H.~Dai, F.~Permenter,
  T.~Koolen, P.~Marion, and R.~Tedrake, ``{Optimization-based locomotion
  planning, estimation, and control design for the atlas humanoid robot},''
  \emph{Auton. Robots}, 2016.

\bibitem{koolen_ijhr}
\BIBentryALTinterwordspacing
T.~Koolen, S.~Bertrand, G.~Thomas, T.~de~Boer, T.~Wu, J.~Smith, J.~Englsberger,
  and J.~Pratt, ``Design of a momentum-based control framework and application
  to the humanoid robot atlas,'' \emph{International Journal of Humanoid
  Robotics}, vol.~13, no.~01, p. 1650007, 2016. [Online]. Available:
  \url{https://doi.org/10.1142/S0219843616500079}
\BIBentrySTDinterwordspacing

\bibitem{Romano2018}
F.~Romano, G.~Nava, M.~Azad, J.~Camernik, S.~Dafarra, O.~Dermy, C.~Latella,
  M.~Lazzaroni, R.~Lober, M.~Lorenzini, D.~Pucci, O.~Sigaud, S.~Traversaro,
  J.~Babic, S.~Ivaldi, M.~Mistry, V.~Padois, and F.~Nori, ``{The CoDyCo Project
  Achievements and Beyond: Toward Human Aware Whole-Body Controllers for
  Physical Human Robot Interaction},'' \emph{IEEE Robot. Autom. Lett.}, vol.~3,
  no.~1, pp. 516--523, jan 2018.

\bibitem{8625025}
G.~Romualdi, S.~Dafarra, Y.~Hu, and D.~Pucci, ``{A Benchmarking of DCM Based
  Architectures for Position and Velocity Controlled Walking of Humanoid
  Robots},'' in \emph{2018 IEEE-RAS 18th Int. Conf. Humanoid Robot.}, 2018, pp.
  1--9.

\bibitem{Natale2017}
L.~Natale, C.~Bartolozzi, D.~Pucci, A.~Wykowska, and G.~Metta, ``{iCub: The
  not-yet-finished story of building a robot child},'' \emph{Sci. Robot.},
  2017.

\bibitem{Marsden2010}
J.~E. Marsden and T.~S. Ratiu, \emph{{Introduction to Mechanics and Symmetry: A
  Basic Exposition of Classical Mechanical Systems}}.\hskip 1em plus 0.5em
  minus 0.4em\relax Springer Publishing Company, Incorporated, 2010.

\bibitem{Murray1994}
R.~M. Murray, S.~S. Sastry, and L.~Zexiang, \emph{{A Mathematical Introduction
  to Robotic Manipulation}}, 1st~ed.\hskip 1em plus 0.5em minus 0.4em\relax
  Boca Raton, FL, USA: CRC Press, Inc., 1994.

\bibitem{Featherstone2014}
R.~Featherstone, \emph{{Rigid Body Dynamics Algorithms}}.\hskip 1em plus 0.5em
  minus 0.4em\relax Springer, Boston, MA, 2014.

\bibitem{Nori2015}
F.~Nori, S.~Traversaro, J.~Eljaik, F.~Romano, A.~{Del Prete}, and D.~Pucci,
  ``{iCub Whole-Body Control through Force Regulation on Rigid Non-Coplanar
  Contacts},'' \emph{Front. Robot. AI}, vol.~2, 2015.

\bibitem{Orin2013}
D.~Orin, A.~Goswami, and S.-H. Lee, ``{Centroidal dynamics of a humanoid
  robot},'' \emph{Auton. Robots}, 2013.

\bibitem{Englsberger2014}
J.~Englsberger, T.~Koolen, S.~Bertrand, J.~Pratt, C.~Ott, and
  A.~Albu-Sch{\"{a}}ffer, ``{Trajectory generation for continuous leg forces
  during double support and heel-to-toe shift based on divergent component of
  motion},'' in \emph{IEEE Int. Conf. Intell. Robot. Syst.}, 2014, pp.
  4022--4029.

\bibitem{Englsberger2019}
J.~Englsberger, G.~Mesesan, C.~Ott, and A.~Albu-Schaffer, ``{DCM-Based Gait
  Generation for Walking on Moving Support Surfaces},'' 2019.

\bibitem{Krause2012}
M.~Krause, J.~Englsberger, P.~B. Wieber, and C.~Ott, ``{Stabilization of the
  Capture Point dynamics for bipedal walking based on model predictive
  control},'' in \emph{IFAC Proc. Vol.}, 2012, pp. 165--171.

\bibitem{Vukobratov2004}
M.~Vukobratov and B.~Borovac, ``{Zero - Moment Point — Thirty Five Years of
  Its Life},'' \emph{Int. J. Humanoid Robot.}, vol.~1, no.~1, pp. 157--173,
  2004.

\bibitem{Choi2007}
Y.~Choi, D.~Kim, Y.~Oh, and B.-j.~J. You, ``{On the Walking Control for
  Humanoid Robot Based on Kinematic Resolution of CoM Jacobian With Embedd ed
  Motion},'' \emph{Proc. 2006 IEEE Int. Conf. Robot. Autom.}, vol.~23, no.~6,
  pp. 1285--1293, 2007.

\bibitem{Olfati-Saber:2001:NCU:935467}
R.~Olfati-Saber, ``{Nonlinear Control of Underactuated Mechanical Systems with
  Application to Robotics and Aerospace Vehicles},'' Ph.D. dissertation,
  Cambridge, MA, USA, 2001.

\bibitem{Metta2010}
G.~Metta, L.~Natale, F.~Nori, G.~Sandini, D.~Vernon, L.~Fadiga, C.~von Hofsten,
  K.~Rosander, M.~Lopes, J.~Santos-Victor, A.~Bernardino, and L.~Montesano,
  ``{The iCub humanoid robot: An open-systems platform for research in
  cognitive development},'' \emph{Neural Networks}, 2010.

\bibitem{osqp}
B.~Stellato, G.~Banjac, P.~Goulart, A.~Bemporad, and S.~Boyd, ``{{\{}OSQP{\}}:
  An Operator Splitting Solver for Quadratic Programs},'' \emph{ArXiv
  e-prints}, 2017.

\bibitem{Torricelli}
D.~{Torricelli}, J.~{Gonzalez-Vargas}, J.~F. {Veneman}, K.~{Mombaur},
  N.~{Tsagarakis}, A.~J. {del-Ama}, A.~{Gil-Agudo}, J.~C. {Moreno}, and J.~L.
  {Pons}, ``Benchmarking bipedal locomotion: A unified scheme for humanoids,
  wearable robots, and humans,'' \emph{IEEE Robotics Automation Magazine},
  vol.~22, no.~3, pp. 103--115, Sep. 2015.

\bibitem{koenig2004design}
N.~Koenig and A.~Howard, ``Design and use paradigms for gazebo, an open-source
  multi-robot simulator,'' in \emph{2004 IEEE/RSJ International Conference on
  Intelligent Robots and Systems (IROS)(IEEE Cat. No. 04CH37566)},
  vol.~3.\hskip 1em plus 0.5em minus 0.4em\relax IEEE, 2004, pp. 2149--2154.

\bibitem{Englsberger2011}
J.~Englsberger, C.~Ott, M.~A. Roa, A.~Albu-Sch{\"{a}}ffer, and G.~Hirzinger,
  ``{Bipedal walking control based on capture point dynamics},'' in \emph{IEEE
  Int. Conf. Intell. Robot. Syst.}, 2011, pp. 4420--4427.

\bibitem{russo}
M.~Russo, D.~Cafolla, and M.~Ceccarelli, ``Design and experiments of a novel
  humanoid robot with parallel architectures,'' \emph{Robotics}, vol.~7, no.~4,
  2018.

\bibitem{Collins}
S.~H. {Collins} and A.~{Ruina}, ``A bipedal walking robot with efficient and
  human-like gait,'' in \emph{Proceedings of the 2005 IEEE International
  Conference on Robotics and Automation}, April 2005, pp. 1983--1988.

\bibitem{Zhang2015}
H.~Zhang, S.~Ahmad, and G.~Liu, ``{Torque Estimation for Robotic Joint With
  Harmonic Drive Transmission Based on Position Measurements},'' \emph{IEEE
  Trans. Robot.}, 2015.

\bibitem{Hopkins_ijhr}
\BIBentryALTinterwordspacing
M.~A. Hopkins, A.~Leonessa, B.~Y. Lattimer, and D.~W. Hong,
  ``Optimization-based whole-body control of a series elastic humanoid robot,''
  \emph{International Journal of Humanoid Robotics}, vol.~13, no.~01, p.
  1550034, 2016. [Online]. Available:
  \url{https://doi.org/10.1142/S0219843615500346}
\BIBentrySTDinterwordspacing

\bibitem{Griffin2016}
R.~J. Griffin and A.~Leonessa, ``{Model predictive control for dynamic footstep
  adjustment using the divergent component of motion},'' in \emph{Proc. - IEEE
  Int. Conf. Robot. Autom.}, 2016.

\bibitem{Shafiee-Ashtiani2017}
M.~Shafiee-Ashtiani, A.~Yousefi-Koma, and M.~Shariat-Panahi, ``{Robust bipedal
  locomotion control based on model predictive control and divergent component
  of motion},'' in \emph{Proc. - IEEE Int. Conf. Robot. Autom.}\hskip 1em plus
  0.5em minus 0.4em\relax IEEE, may 2017, pp. 3505--3510.

\bibitem{shafiee2019online}
M.~Shafiee, G.~Romualdi, S.~Dafarra, F.~J.~A. Chavez, and D.~Pucci, ``{Online
  DCM Trajectory Generation for Push Recovery of Torque-Controlled Humanoid
  Robots},'' in \emph{2019 IEEE-RAS 19th Int. Conf. Humanoid Robot.}, 2019.

\end{thebibliography}
\bibliographystyle{bib_template}

\vspace*{8pt}  
\noindent%
\parbox{5truein}{
\begin{minipage}[b]{1truein}
\centerline{{\psfig{file=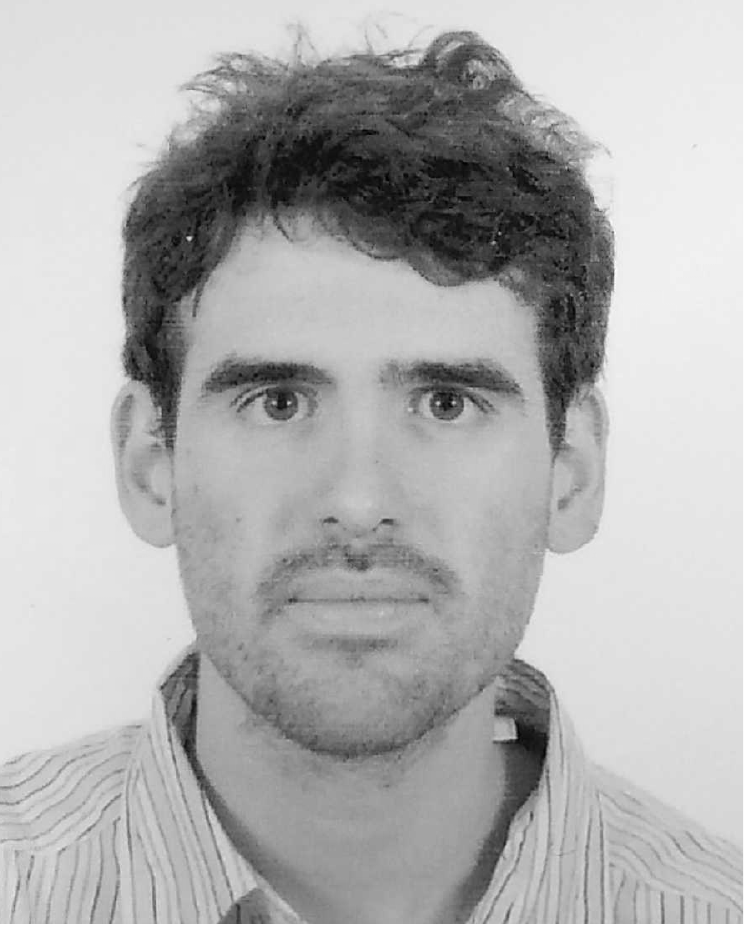,width=1in,height=1.25in}}}
\end{minipage}
\hfill %
\begin{minipage}[b]{3.85truein}
{{\bf Giulio Romualdi} received his B.S. degree in Biomedical Engineering from the University of Pisa, Italy, and his M.S. degree in Robotics and Automation Engineering from the University of Pisa,
Italy, in 2014 and 2018, respectively. He is currently a Ph.D. student at the University of Genova, Italy and Istituto Italiano di Tecnologia, Italy. His research interests include bipedal locomotion, humanoid robots and nonlinear control theory.}
\end{minipage} } %

\vspace*{8pt}  
\noindent%
\parbox{5truein}{
\begin{minipage}[b]{1truein}
\centerline{{\psfig{file=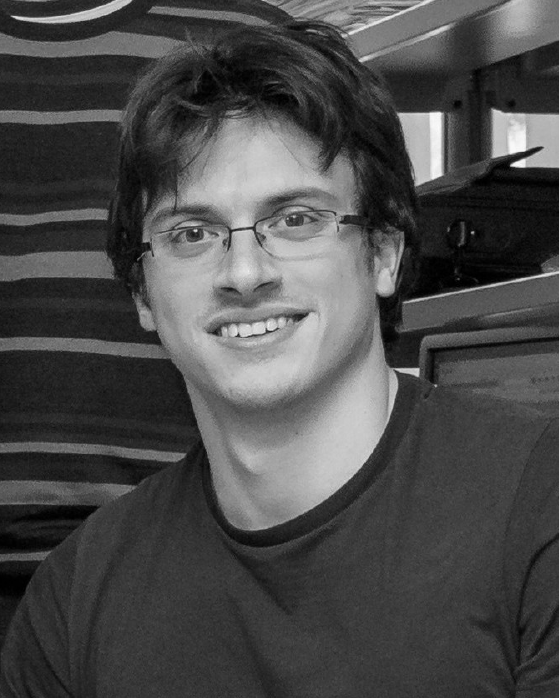,width=1in,height=1.25in}}}
\end{minipage}
\hfill %
\begin{minipage}[b]{3.85truein}
{{\bf Stefano Dafarra} received his M.S. degree on Automation and Control Engineering from Politecnico di Milano in 2016. He is currently a Ph.D. candidate on Advanced and Humanoid Robotics at the Istituto Italiano di Tecnologia. His research interest includes optimization, optimal control, and humanoid locomotion.}\end{minipage} } 

\vspace*{8pt}  
\noindent%
\parbox{5truein}{
\begin{minipage}[b]{1truein}
\centerline{{\psfig{file=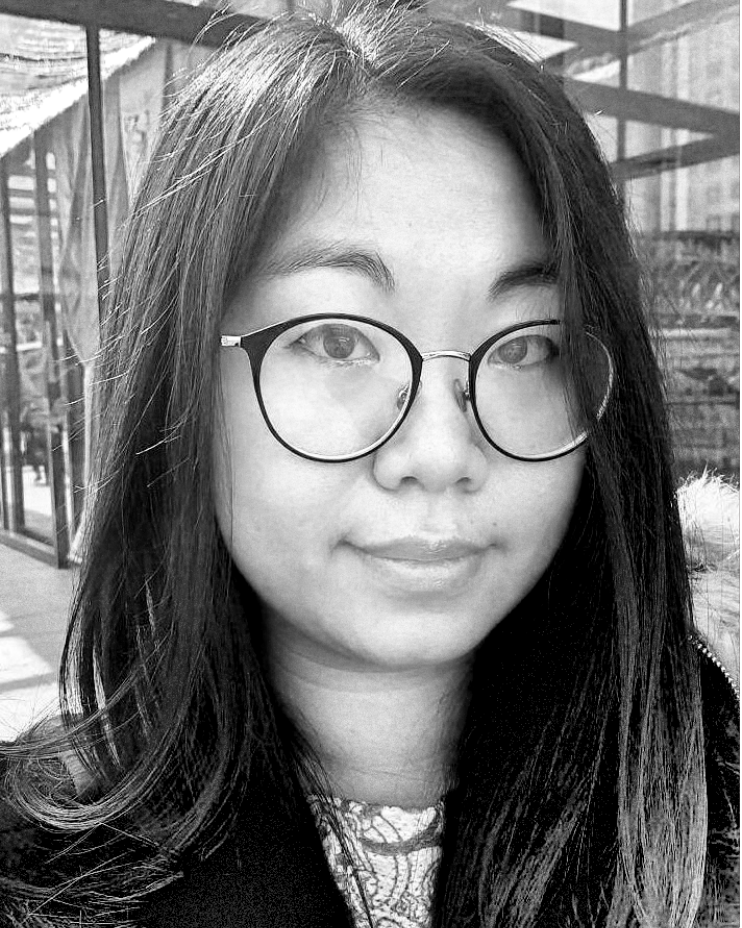,width=1in,height=1.25in}}}
\end{minipage}
\hfill %
\begin{minipage}[b]{3.85truein}
{{\bf Yue Hu} received her PhD degree in robotics from Heidelberg University in 2017. She was postdoc first at Heidelberg University in Germany, then at Fondazione Istituto Italiano di Tecnologia (IIT) in Italy. She is now a JSPS (Japan Society for the Promotion of Science) fellow at the National Institute of Advanced Industrial Science and Technologies (AIST) in Japan, as member of the CNRS-AIST JRL (Joint Robotics Laboratory),}\end{minipage} }
\vskip 2pt
\noindent
UMI3218/RL. Her research interests focus mainly on optimal control, humanoid robots, bipedal locomotion, human motion analysis, and physical human-robot interaction. 

\vspace*{8pt}  
\noindent%
\parbox{5truein}{
\begin{minipage}[b]{1truein}
\centerline{{\psfig{file=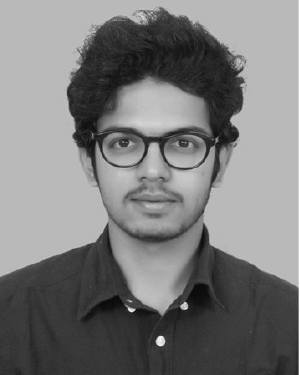,width=1in,height=1.25in}}}
\end{minipage}
\hfill %
\begin{minipage}[b]{3.85truein}
{{\bf Prashanth Ramadoss} received his double M.S.c/M.E. degree, Masters in Robotics Engineering, M.E., from the University of Genova, Italy (M2) and a Research Master's degree in Advanced Robotics, M.Sc. from Ecole Centrales de Nantes, France (M1), in 2015. He is currently a PhD student in University of Genova, Italy and Istituto Italiano di Tecnologia, Italy. His research interests include multi-body systems, contact dynamics, geometric}\end{minipage}}
\vskip 2pt
\noindent
observers, sensor fusion, estimation and control. 

\vspace*{8pt}  
\noindent%
\parbox{5truein}{
\begin{minipage}[b]{1truein}
\centerline{{\psfig{file=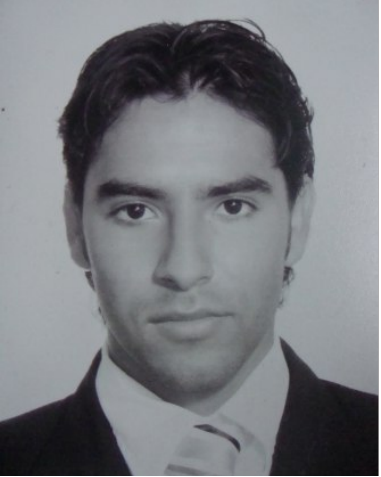,width=1in,height=1.25in}}}
\end{minipage}
\hfill %
\begin{minipage}[b]{3.85truein}
{{\bf Francisco Javier Andrade Chavez} is currently a PostDoc in the Dynamic Interaction Control Lab at Istituto Italiano di Tecnologia. He received a double degree Master in Advance Robotics from Unversita degli Studi di Genova and Ecole Central de Nantes in 2015. He has recently successfully concluded a PhD in Bioengineering and Robotics at Universita degli Studi di Genova in cooperation with the Istituto Italiano di Tecnologia}\end{minipage}}
\vskip 2pt
\noindent
under the supervision of Daniele Pucci. His main research interests are sensing, estimation ,and control of dynamics and its application to bio-inspired robots and human-robot collaboration.

\vspace*{8pt}  
\noindent%
\parbox{5truein}{
\begin{minipage}[b]{1truein}
\centerline{{\psfig{file=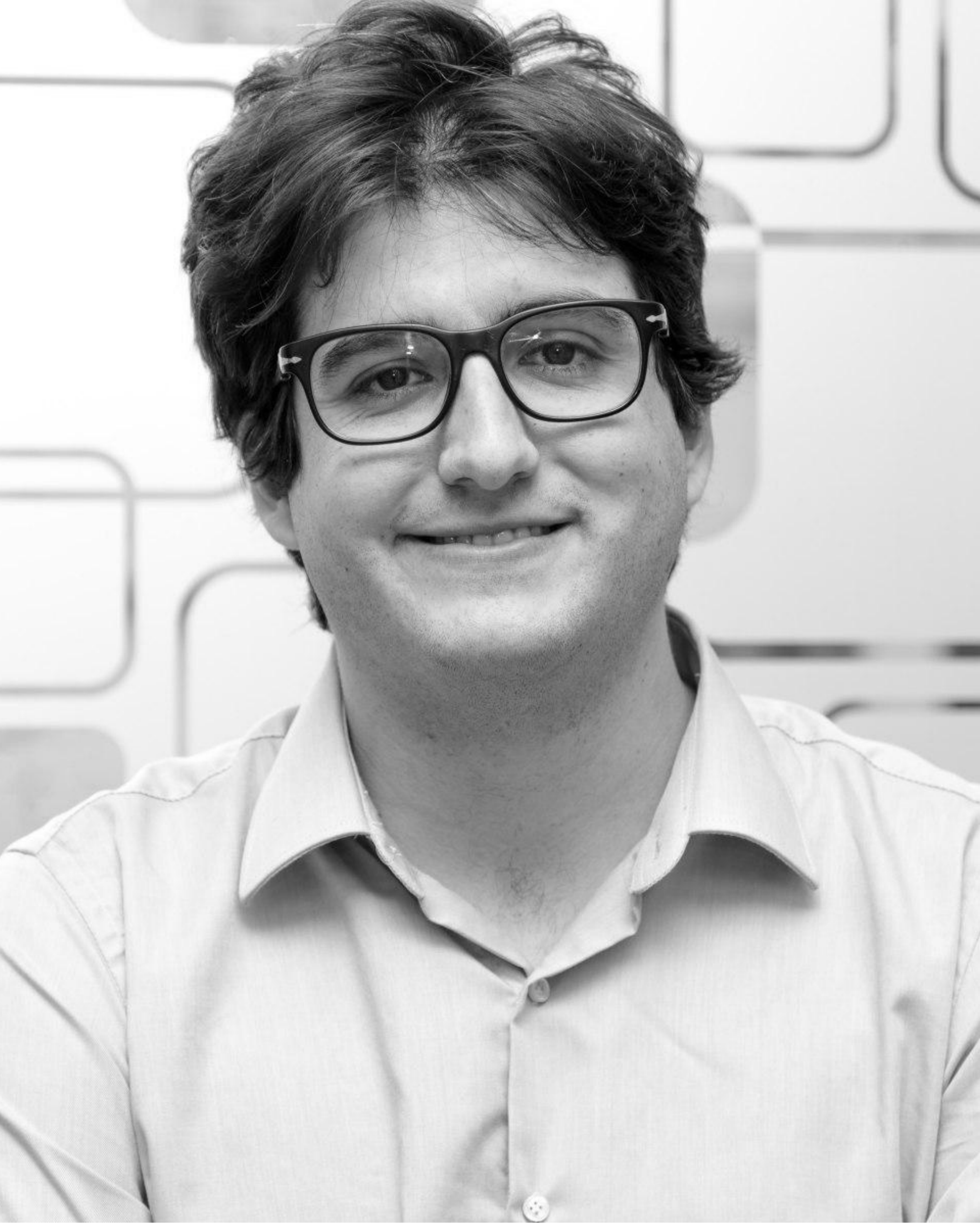,width=1in,height=1.25in}}}
\end{minipage}
\hfill %
\begin{minipage}[b]{3.85truein}
{{\bf Silvio Traversaro} received the BEng and MEng degrees in Computer Science and Robotics Engineering with highest honors from the University of Genova, in 2011 and 2013, respectively. In 2017 he received the Ph.D. in Robotics from the University of Genova working at the Istituto Italiano di Tecnologia on dynamics modelling, estimation and identification applied to the iCub robot. Since then, he has been a post-doctoral researcher \hfilneg}\end{minipage}} 
\vskip 2pt
\noindent
at the Istituto Italiano di Tecnologia, working on industrial and commercial applications of robotics.

\vspace*{8pt}  
\noindent%
\parbox{5truein}{
\begin{minipage}[b]{1truein}
\centerline{{\psfig{file=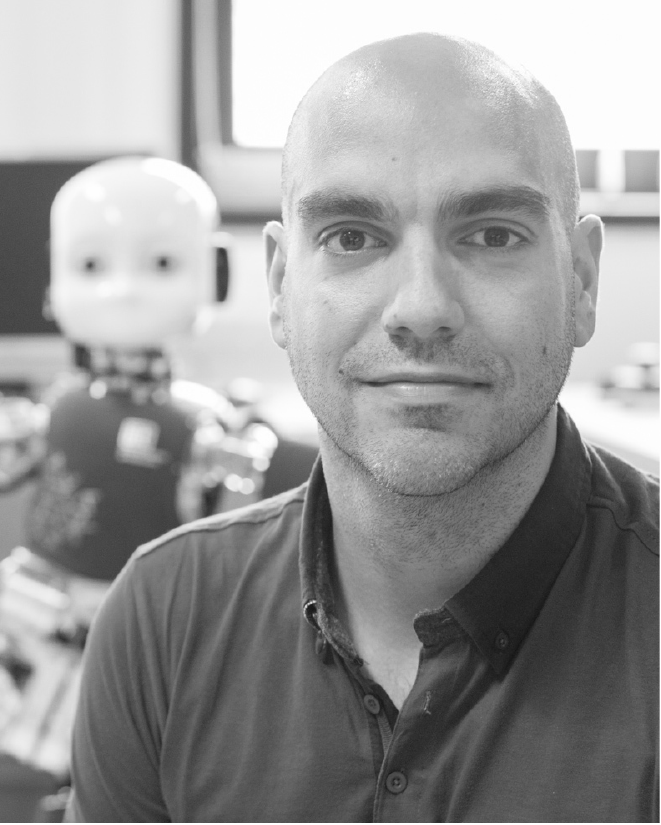,width=1in,height=1.25in}}}
\end{minipage}
\hfill %
\begin{minipage}[b]{3.85truein}
{{\bf Daniele Pucci} received the bachelor and master degrees in Control Engineering with highest honours from ”Sapienza”, University of Rome. In 2009, ”Sapienza”, University of Rome, also gave him the ”Academic Excellence Award” along with the master. In 2013, he earned the PhD title with a thesis prepared at INRIA Sophia Antipolis, France, under the supervision of Tarek Hamel and Claude Samson. From 2013 to 2017, he has been a postdoc \hfilneg}\end{minipage}} 
\vskip 1pt  
\noindent
 at the Istituto Italiano di Tecnologia (IIT) working within the EU project CoDyCo. Since August 2017, he is the head of the Dynamic Interaction Control (DIC) lab and the PI of the H2020 European Project AnDy. He is also the PI of three joint labs with Camozzi Automation, Danieli Automation, and Honda Research Japan. The DIC lab research focus spans three research axes: Telexistence, Human-Robot Collaboration, and Aerial Humanoid Robotics. In particular, Aerial Humanoid Robotics is pioneering the feasibility and application of flying humanoid robots.

\vfill\eject

\end{document}